\documentclass{article}

\usepackage{iclr2026_conference,times}
\iclrfinalcopy
\usepackage{hyperref}
\usepackage{url}
\usepackage{booktabs}
\usepackage{amsmath}
\usepackage{amssymb}
\usepackage{graphicx}
\usepackage{multirow}
\usepackage{xcolor}
\usepackage{enumitem}
\usepackage{wrapfig}
\usepackage{float}
\usepackage{tabularx}
\hypersetup{hidelinks,hypertexnames=false}

\newcolumntype{Y}{>{\centering\arraybackslash}X}

\usepackage{amsmath,amsfonts,bm}

\def\eqref#1{equation~\ref{#1}}

\def\1{\bm{1}}

\DeclareMathAlphabet{\mathsfit}{\encodingdefault}{\sfdefault}{m}{sl}
\SetMathAlphabet{\mathsfit}{bold}{\encodingdefault}{\sfdefault}{bx}{n}

\newcommand{\method}{Self Gradient Forcing}
\newcommand{\methodshort}{SGF}
\newcommand{\kv}{KV}

\newcommand{\sfcell}{SF}
\newcounter{algorithm}
\renewcommand{\thealgorithm}{\arabic{algorithm}}

\graphicspath{{figures/}}

\title{Self Gradient Forcing: Native Long Video Extrapolation}

\author{
Junhao Zhuang \quad Shiyi Zhang \quad Yuxuan Bian \quad Yaowei Li\\
Yawen luo \quad Yijun Liu \quad Weiyang Jin \quad Songchun Zhang \quad Xianglong He\\
Xuying Zhang \quad Haoran Li \quad Haoyang Huang \quad Zeyue Xue \quad Nan Duan\\
Joy Future Academy, JD
}

\makeatletter
\def\@maketitle{\vbox{\hsize\textwidth
{\LARGE\sc \@title\par}
\ificlrfinal
    \lhead{}
    \def\And{\end{tabular}\hfil\linebreak[0]\hfil
            \begin{tabular}[t]{c}\rule{\z@}{24pt}\ignorespaces}%
    \def\AND{\end{tabular}\hfil\linebreak[4]\hfil
            \begin{tabular}[t]{c}\rule{\z@}{24pt}\ignorespaces}%
    \begin{center}
    \begin{tabular}[t]{c}\rule{\z@}{24pt}\@author\end{tabular}%
    \end{center}
\else
    \lhead{}
    \def\And{\end{tabular}\hfil\linebreak[0]\hfil
            \begin{tabular}[t]{c}\rule{\z@}{24pt}\ignorespaces}%
    \def\AND{\end{tabular}\hfil\linebreak[4]\hfil
            \begin{tabular}[t]{c}\rule{\z@}{24pt}\ignorespaces}%
    \begin{center}
    \begin{tabular}[t]{c}\rule{\z@}{24pt}\@author\end{tabular}%
    \end{center}
\fi
\vskip 0.3in minus 0.1in}}
\makeatother

\begin{document}

\maketitle

\begin{abstract}
Recent autoregressive video diffusion methods are increasingly built upon Self Forcing, where the student is trained on histories produced by its own rollout rather than ground-truth video contexts.
This reduces exposure bias, but the historical key-value cache is still used by future frames only as frozen rollout state. As a result, future losses cannot supervise how earlier generated latents should be written into more useful keys and values for later video-latent generation. 
We call this the historical context-gradient gap. We propose Self Gradient Forcing (SGF), a two-pass training strategy that restores this missing supervision signal without backpropagating through the full serial rollout. 
Pass 1 performs a no-gradient autoregressive rollout matching inference and, at a sampled denoising exit step, records both the self-generated context and the noisy latents fed to the model. 
Pass 2 performs parallel context-gradient reconstruction for the recorded exit step. 
The generated context is used as stop-gradient clean-latent input, while the model recomputes the context KV representations and future-to-context causal attention. 
Thus, SGF provides the missing memory-writing supervision within the native autoregressive training objective, using losses on future video latents to train the model to encode context into more effective causal memory. 
Across extensive long-horizon frame-wise and chunk-wise experiments under different initializations, SGF achieves stronger native long-video extrapolation than Self Forcing, especially in subject identity, background/layout consistency, and temporal stability. 
Remarkably, using only a 5-second training window, SGF can extrapolate to videos lasting several minutes.
Code and models will be released on the project page \url{https://zhuang2002.github.io/SelfGradientForcing}.
\end{abstract}

\section{Introduction}

Long-form autoregressive video generation requires scenes, objects, layouts, and dynamics to remain coherent as generation extends far beyond the training window. In autoregressive video diffusion, this coherence depends on extrapolating from the model's generated history: once rollout begins, each new frame or chunk is conditioned on the prompt and previously generated content.

However, in Teacher-Forcing (TF) training, the model is conditioned on ground-truth video contexts, whereas at inference it is conditioned on self-generated histories.
This training--inference mismatch creates exposure bias and makes native long-video extrapolation difficult.
Self Forcing~\citep{selfforcing} mitigates the mismatch by training the student on histories produced by its own autoregressive rollout, using distribution-matching distillation (DMD) supervision from a bidirectional video model.
This self-rollout recipe has become an increasingly common training paradigm for autoregressive video diffusion~\citep{selfforcing,rollingforcing,selfforcingpp,longlive,longlive2,contextforcing,sparseforcing,forcingkv,headforcing,longliverag,tethercache}.
However, to make self-rollout training feasible, these methods typically inherit a crucial computational constraint: cross-chunk gradient flow through the historical \kv{} cache is truncated.

\begin{figure}[t]
    \includegraphics[width=1.0\textwidth]{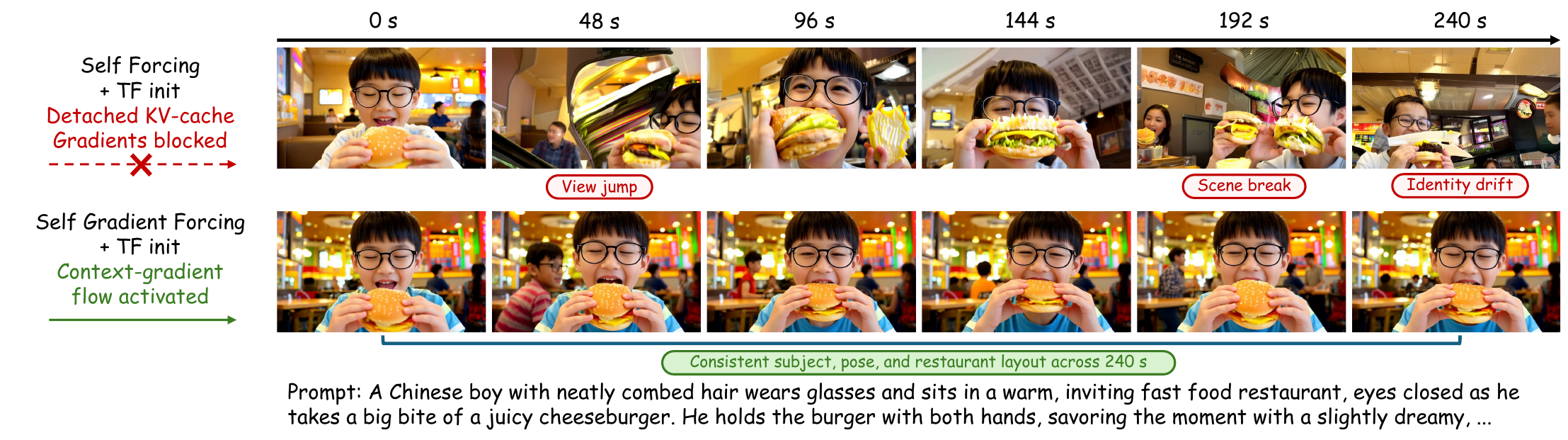}
    \vspace{-6mm}
    \caption{
    \textbf{Long-horizon consistency with Self Gradient Forcing.}
    Under the same prompt, seed, and TF initialization, Self Forcing trains with self-generated history but does not supervise the clean-timestep \kv{} writing computation with future losses; it eventually exhibits view jumps, scene breaks, and identity drift. SGF restores a bounded context-gradient path for self-generated memory writing and better preserves subject identity, pose, and restaurant layout over 240 seconds.
    }
    \vspace{-4mm}
    \label{fig:teaser} 
\end{figure}

The consequence is that Self Forcing exposes the model to self-generated history, but does not let future losses train how that history should be written as future-readable memory. During rollout, generated latents are processed by the causal DiT at the clean context timestep and stored as a causal \kv{} cache; later chunks read this cache as frozen historical context. Future losses can therefore train noisy denoising tokens to read cached history, but cannot propagate into the clean-timestep computation that wrote the historical K/V entries. We call this missing credit-assignment path the \emph{historical context-gradient gap}.
Fig.~\ref{fig:teaser} illustrates the qualitative effect: a Self-Forcing model can remain locally plausible for many frames, yet identity, viewpoint, and layout consistency gradually break down as extrapolation continues.

After TF, consistency-distillation (CD), or ODE initialization~\citep{selfforcing,causalforcing,causalforcingpp}, the model has a useful prior for writing real video contexts into the cache at the clean context timestep.  
However, the final self-rollout objective changes both the context distribution and the supervision: history is self-generated, and few-step DMD losses are applied at sampled noisy denoising timesteps.  
This leaves a cache-writing gap for self-generated histories.  
The gap can further widen because the same causal DiT shares parameters across timesteps: updates from noisy denoising steps can alter the clean $t_{\mathrm{ctx}}=0$ cache-writing computation, while later-chunk losses do not backpropagate to the historical \kv{} entries it wrote.  
The context-writing path may therefore drift from what long autoregressive rollout requires.

A direct solution would be to keep the historical \kv{} cache differentiable, allowing future losses to backpropagate into the clean-timestep K/V writing computation.
In practice, this would require retaining the autograd graph for every historical cache write until future chunks consume it.
These graphs grow with rollout length, transformer depth, and sequential cache updates, making the direct KV-gradient path difficult to scale.

We propose \method{} (\methodshort{}), which turns this serial graph-retention problem into a bounded parallel recomputation problem.
SGF is a two-pass training strategy that restores the missing memory-writing supervision without full backpropagation through the sampled rollout.
Pass 1 performs the true serial no-gradient autoregressive rollout and records the self-generated context together with the noisy latents fed to the model at a sampled denoising exit step.
Pass 2 discards the rollout cache and reconstructs the same exit-step computation in parallel: the generated context is reprocessed as stop-gradient clean-latent input, and the recorded noisy latents are fed again as prediction inputs.
The model recomputes the context hidden states, \kv{} representations, and future-to-context causal attention, so losses on future video latents update the shared parameters that write clean-timestep K/V representations for later generation.

As a native training framework for forcing-based autoregressive video diffusion, SGF recovers the context-writing gradient unused by frozen-cache Self Forcing and is orthogonal to existing forcing improvements, allowing it to be applied directly on top of them.
We evaluate SGF across multiple initializations and both frame-wise and chunk-wise autoregressive generation.
Empirically, SGF is comparable to Self Forcing at 5 seconds and substantially improves native long-video extrapolation at 60 and 240 seconds, with the clearest gains in subject identity, background/layout consistency, and temporal stability.
Notably, models trained with only a 5-second window can still extrapolate to minute-scale videos.
We further analyze two-pass recovery accuracy, the start-boundary effect of causal video VAEs, and how the number of sink latents affects streaming generation quality.

Our contributions are:
\begin{itemize}[leftmargin=*,topsep=2pt,itemsep=1pt,parsep=0pt]
    \item We identify the historical context-gradient gap in frozen-cache Self Forcing, where future losses supervise cache reading but not the clean context-writing computation that writes self-generated history into K/V memory; as shared DiT parameters are updated at other timesteps during few-step DMD, this unsupervised path can drift.
    \item We introduce \methodshort{}, a two-pass training strategy whose serial Pass 1 records the self-generated rollout state and whose parallel Pass 2 performs context-gradient reconstruction of the sampled exit-step computation with gradients through self-generated context \kv{} representations and future-to-context causal attention.
    \item We conduct extensive experiments on frame-wise and chunk-wise generation under different initializations at 5s, 60s, and 240s, showing that SGF matches Self Forcing at 5s and substantially improves long-video extrapolation at 60s and 240s. We study two-pass recovery accuracy, causal VAE boundary diagnostics, and the effect of sink latents on streaming generation quality.
\end{itemize}
\section{Related Work}

\paragraph{Autoregressive video diffusion.}
Modern video generation builds on diffusion objectives, accelerated samplers, latent diffusion, and diffusion transformers, which provide the modeling and scaling principles used by image and video generators~\citep{ddpm,ddim,latentdiffusion,dit}. Video diffusion systems extend these foundations with 3D denoising, cascaded generation, latent video modeling, motion modules, space-time architectures, and multimodal autoregressive token modeling~\citep{videodiffusionmodels,imagenvideo,makeavideo,videoldm,stablevideodiffusion,animatediff,videocrafter2,lumiere,videopoet}. For long-form generation, however, the key distinction is the causal use of generated history: each generated latent frame or chunk becomes part of the context for later generation. Diffusion Forcing~\citep{diffusionforcing} connects next-token prediction and diffusion by independently noising sequence elements, and CausVid~\citep{causvid} distills bidirectional video diffusion into a fast autoregressive student with causal caching. SGF follows this autoregressive video diffusion interface and asks whether self-generated history, once written into causal \kv{} state, receives future supervision as future-readable memory.

\paragraph{Forcing objectives for self-generated histories.}
Teacher-forcing trains on ground-truth prefixes, whereas autoregressive inference conditions on self-generated histories. Self Forcing~\citep{selfforcing} directly reduces this exposure bias by training the student on histories produced by its own rollout, using distribution-matching distillation supervision from a bidirectional teacher. Subsequent forcing methods build on this self-rollout training paradigm. Causal Forcing~\citep{causalforcing} and Causal Forcing++~\citep{causalforcingpp} address a complementary initialization mismatch: a causal student should not depend on bidirectional teacher trajectories whose flow map is unavailable at inference. Rolling Forcing~\citep{rollingforcing}, Self-Forcing++~\citep{selfforcingpp}, Matrixgame3~\citep{matrixgame3}, and ShotStream~\citep{shotstream} further extend rollout exposure through longer segments and windowed denoising. Video-Mirai~\citep{videomirai} and Next Forcing~\citep{nextforcing} make a related observation that next-step supervision can discard information needed by later frames, and introduce future-aware or multi-chunk supervision. These methods improve the history distribution, causal initialization, or temporal reach of the supervision. SGF addresses a remaining gap in frozen-cache self-rollout training: future losses can supervise how later noisy denoising tokens read cached history, but not how the clean-timestep computation writes self-generated history into \kv{} memory.

\paragraph{Long-horizon context and cache design.}
A line of work improves long-horizon generation by changing which context is available, how it is positioned, or how it is retained. Gen-L-Video performs overlapping temporal co-denoising for long multi-text videos~\citep{genlvideo}; FreeNoise reschedules noise and fuses windowed temporal attention~\citep{freenoise}; FIFO-Diffusion maintains a frame queue at different noise levels~\citep{fifodiffusion}; and StreamingT2V combines short-term and long-term memory in an autoregressive pipeline~\citep{streamingt2v}. Recurrent sequence modeling and efficient attention study segment recurrence~\citep{transformerxl}, rotary position embeddings~\citep{rope}, attention sinks~\citep{streamingllm}, cache retention~\citep{h2o,snapkv}, streaming long tuning~\citep{longlive,longlive2}, long-context supervision~\citep{contextforcing}, trainable sparse attention~\citep{sparseforcing,flashvsr}, KV compression~\citep{forcingkv}, head-wise cache behavior~\citep{headforcing}, retrieval-augmented latent history~\citep{longliverag}, and gated recall with trusted alignment~\citep{tethercache}. These methods change the memory exposed to the generator. SGF is orthogonal to this direction: for a given context and cache design, it improves how self-generated content is written into future-readable \kv{} representations. We evaluate SGF across frame-wise and chunk-wise generation, multiple initializations, and extrapolation horizons, showing that better \kv{} memory writing improves long-horizon autoregressive video generation even under short-window forcing training.

\section{Method}

\subsection{Historical Context-Gradient Gap}

Autoregressive video diffusion generates latent blocks sequentially.
At block $j$, the causal generator denoises $z_j^t$ while attending to a historical K/V cache, rather than to raw past latents.
After block $i<j$ is generated, its predicted clean latent $\tilde{x}_i$ is processed at the clean context timestep $t_{\mathrm{ctx}}=0$, and the resulting K/V entries are appended to the cache.
Thus, later blocks condition on the clean-timestep memory representation written from $\tilde{x}_i$, not on $\tilde{x}_i$ directly.

Let $\mathcal{C}_\theta$ denote the cache-writing computation induced by the causal generator at the clean context timestep.
In a serial rollout, this update is recurrent: the model reads the existing cache and writes the next historical K/V entry,
\begin{equation}
\begin{aligned}
\mathsf{KV}_i^0(\theta)
=
\mathcal{C}_\theta\!\left(
\tilde{x}_i,t_{\mathrm{ctx}};\mathsf{KV}_{<i}^0
\right),
\qquad
t_{\mathrm{ctx}}=0.
\end{aligned}
\end{equation}
The new entry $\mathsf{KV}_i^0$ then becomes part of the state used by later blocks, including subsequent cache updates.
If future losses could differentiate through this historical entry, they would assign credit to the clean-context computation that encoded $\tilde{x}_i$ into K/V memory.
This would train not only how later noisy tokens read self-generated history, but also how earlier generated latents should be written into memory for future denoising.

Frozen-cache Self Forcing removes this memory-writing signal.
It trains on self-generated histories, reducing the mismatch between training and inference, and later-block losses can still update the target-side denoising computation that reads the recorded cache.
However, the historical K/V entries are treated as detached rollout state.
Consequently, future losses do not supervise the $t_{\mathrm{ctx}}=0$ computation that produced those entries.

\begin{figure}[t]
\centering
\includegraphics[width=\linewidth]{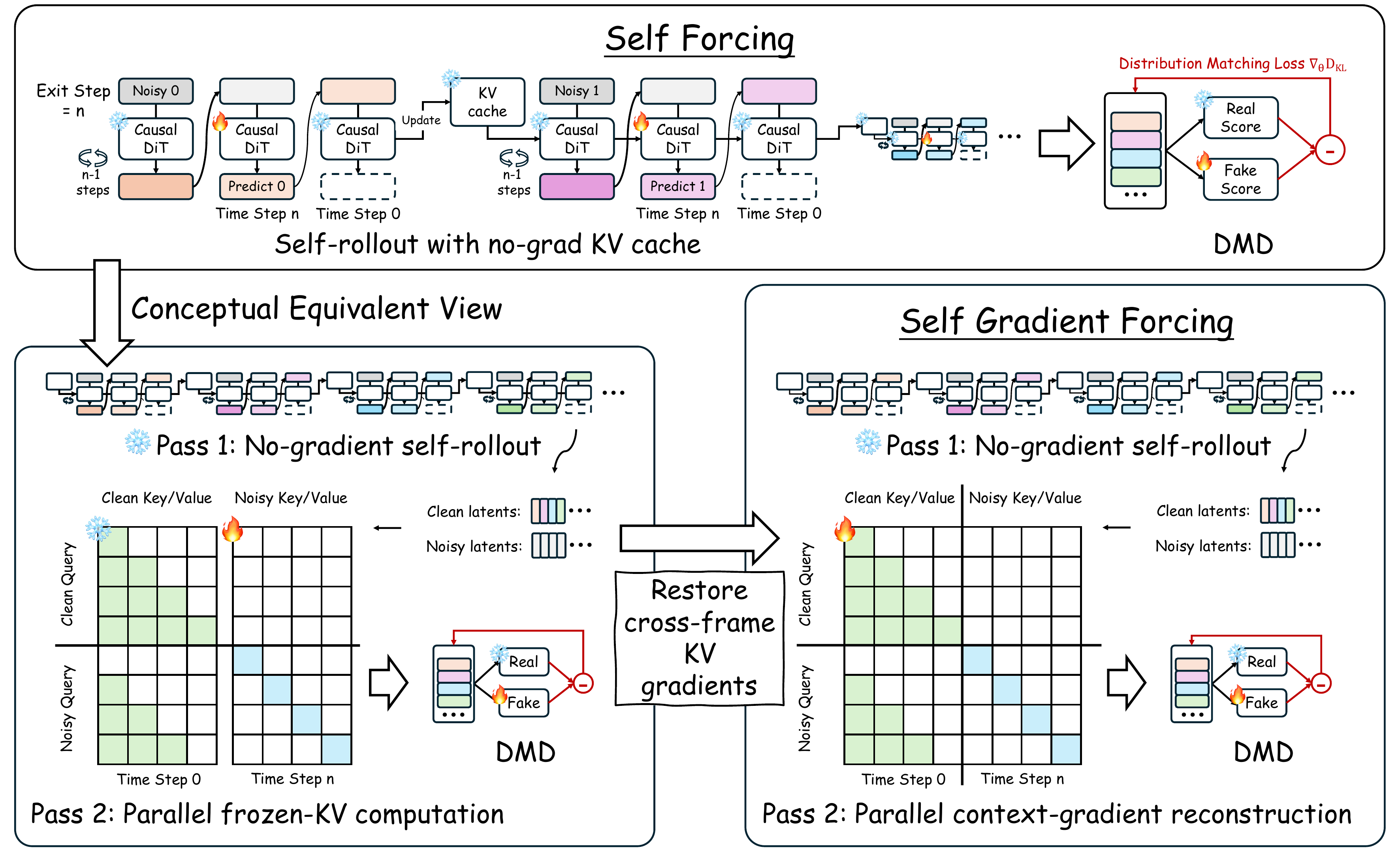}
\vspace{-6mm}
\caption{
\textbf{From frozen-cache Self Forcing to Self Gradient Forcing.}
Self Forcing trains on self-generated histories but treats historical K/V entries as detached cache state.
SGF keeps the no-gradient rollout unchanged and adds a parallel reconstruction pass, where detached context latents are re-encoded at the clean context timestep so future DMD losses supervise K/V writing without serial rollout backpropagation.
}
\vspace{-4mm}
\label{fig:sgf-flowchart}
\end{figure}

This omission becomes problematic during final Self Forcing because the same DiT parameters are shared across noisy denoising and clean-context cache writing.
DMD losses at noisy timesteps update the shared parameters,
\begin{equation}
\begin{aligned}
\theta_{r+1}
=
\theta_r-\eta\nabla_\theta \mathcal{L}_{\mathrm{SF}}(\theta_r),
\qquad
\mathsf{KV}_i^0(\theta_{r+1})
\not\equiv
\mathsf{KV}_i^0(\theta_r)
\quad \text{in general}.
\end{aligned}
\end{equation}
Thus, training can change the clean-context cache writer, but frozen-cache Self Forcing provides no future-loss correction for how self-generated latents are encoded into K/V memory.
We call this missing supervision path the \emph{historical context-gradient gap}.
Figure~\ref{fig:sgf-flowchart} illustrates how SGF restores this missing path by replacing frozen-cache reconstruction with context-gradient reconstruction.

\subsection{Direct Differentiable Cache}

A direct way to close the gap is to keep the serial historical K/V cache differentiable.
Then future losses could backpropagate through the cached K/V entries into the earlier clean-context calls that produced them.
However, this requires retaining the backward graph for every cache write until all later blocks that consume it have been processed.
Because autoregressive rollout repeatedly writes, reads, and extends the cache across denoising steps and transformer layers, the graph grows with rollout length rather than remaining a fixed-window computation.
The bottleneck is therefore not the storage of K/V tensors alone, but the saved activations attached to the recurrent cache trajectory.
This makes direct differentiable-cache training impractical for long-horizon self-rollout.

\subsection{Self Gradient Forcing}
\begin{wrapfigure}[16]{r}{0.55\linewidth}
\vspace{-10pt}
\centering
\refstepcounter{algorithm}
\label{alg:sgf-training}

\begin{minipage}{\linewidth}
\tiny
\renewcommand{\arraystretch}{0.88}
\setlength{\tabcolsep}{0pt}

\noindent\rule{\linewidth}{0.7pt}
\textbf{Algorithm \thealgorithm{} \method{} Training}
\par
\vspace{-2pt}
\noindent\rule{\linewidth}{0.4pt}

\begin{tabular}{@{}r@{\hspace{0.3em}}p{0.90\linewidth}@{}}
\multicolumn{2}{@{}p{\linewidth}@{}}{\textbf{Require:} Denoising steps $\mathcal{T}=(t_1,\ldots,t_K)$ and scheduler $\Psi$}\\
\multicolumn{2}{@{}p{\linewidth}@{}}{\textbf{Require:} Autoregressive blocks $N$ and clean context timestep $t_{\mathrm{ctx}}=0$}\\
\multicolumn{2}{@{}p{\linewidth}@{}}{\textbf{Require:} Reconstruction causal mask $\mathcal{M}_{\mathrm{rec}}$}\\
\multicolumn{2}{@{}p{\linewidth}@{}}{\textbf{Require:} Causal DiT $G_\theta$ with serial cache and parallel reconstruction interfaces}\\

1: & \textbf{loop}\\
2: & \quad Initialize serial cache $\mathsf{KV}\leftarrow []$ and records $Z^\star,X_{\mathrm{ctx}}\leftarrow []$\\
3: & \quad Sample exit index $s\sim\mathrm{Uniform}\{1,\ldots,K\}$ and set $t^\star\leftarrow t_s$\\
4: & \quad \textbf{Pass 1: no-gradient self-rollout}\\
5: & \quad Disable gradient computation for all Pass-1 operations\\
6: & \quad \textbf{for} block $i=1,\ldots,N$ \textbf{do}\\
7: & \quad\quad Sample initial noisy latent $z_i^{t_1}\sim\mathcal{N}(0,I)$\\
8: & \quad\quad \textbf{for} denoising step $k=1,\ldots,s$ \textbf{do}\\
9: & \quad\quad\quad $\hat{x}_i^{(k)}\leftarrow G_\theta(z_i^{t_k};t_k,\mathsf{KV})$\\
10: & \quad\quad\quad \textbf{if} $k<s$ \textbf{then} sample $\epsilon$ and set $z_i^{t_{k+1}}\leftarrow\Psi(\hat{x}_i^{(k)},\epsilon,t_{k+1})$ \textbf{end if}\\
11: & \quad\quad \textbf{end for}\\
12: & \quad\quad Set $Z_i^\star\leftarrow z_i^{t^\star}$ and $\tilde{x}_i\leftarrow\hat{x}_i^{(s)}$\\
13: & \quad\quad $\mathsf{KV}\leftarrow
\mathsf{KV}\cup\mathcal{C}_\theta(\tilde{x}_i,t_{\mathrm{ctx}};\mathsf{KV})$\\
14: & \quad\quad Append $\tilde{x}_i$ to $X_{\mathrm{ctx}}$ and $Z_i^\star$ to $Z^\star$\\
15: & \quad \textbf{end for}\\
16: & \quad \textbf{Pass 2: parallel context-gradient reconstruction}\\
17: & \quad Set $X_{\mathrm{rec}}\leftarrow\operatorname{sg}(X_{\mathrm{ctx}})$\\
18: & \quad Enable gradient computation\\
19: & \quad $\hat{X}_{\mathrm{tar}}\leftarrow
G_\theta(Z^\star,t^\star;X_{\mathrm{rec}},t_{\mathrm{ctx}},\mathcal{M}_{\mathrm{rec}})$\\
20: & \quad Update $\theta$ with $\mathcal{L}_{\mathrm{DMD}}(\hat{X}_{\mathrm{tar}})$; keep context-K/V gradients\\
21: & \textbf{end loop}
\end{tabular}

\par\vspace{2pt}
\noindent\rule{\linewidth}{0.7pt}
\end{minipage}
\vspace{-12pt}
\end{wrapfigure}

Algorithm~\ref{alg:sgf-training} summarizes the SGF training step.
SGF can be understood from an equivalent two-pass view of Self Forcing.
In this view, Pass 1 performs the ordinary no-gradient self-rollout and records a sampled exit state, while Pass 2 reconstructs the corresponding causal computation in parallel.
Frozen-cache Self Forcing treats the reconstructed context K/V path as detached memory in Pass 2.
SGF keeps Pass 1 and the reconstruction geometry unchanged, but changes the Pass-2 gradient boundary so future DMD losses also supervise clean-context K/V writing.

\paragraph{Pass 1: no-gradient self-rollout.}
The first pass is the ordinary serial autoregressive rollout used at inference.
For each block $i$, the model denoises with the current historical cache and records a sampled exit state: the noisy input $z_i^{t^\star}$ and the predicted clean latent $\tilde{x}_i$.
The latent $\tilde{x}_i$ is then processed at the clean context timestep $t_{\mathrm{ctx}}=0$ to update the serial K/V cache for later blocks.
All computations in this pass are run without gradient tracking, and the recorded states are treated as fixed data for Pass 2.

\paragraph{SF Pass 2: frozen-K/V computation.}
Frozen-cache Self Forcing can be viewed as a parallel reconstruction of the sampled exit computation with a detached context K/V path.
Given the recorded noisy latents $Z^\star=\{z_i^{t^\star}\}_{i=1}^N$ and context latents $\tilde{X}_{\mathrm{ctx}}=\{\tilde{x}_i\}_{i=1}^N$, it reconstructs target predictions under the causal mask $\mathcal{M}_{\mathrm{rec}}$:
\begin{equation}
\begin{aligned}
\hat{X}_{\mathrm{tar}}
=
G_\theta\!\left(
Z^\star,t^\star;
\tilde{X}_{\mathrm{ctx}},t_{\mathrm{ctx}},
\mathcal{M}_{\mathrm{rec}}
\right).
\end{aligned}
\end{equation}
The mask $\mathcal{M}_{\mathrm{rec}}$ reproduces the sink-plus-window attention relation induced by the serial cache.
In frozen-cache Self Forcing, however, the K/V entries produced by the clean-context side are treated as detached memory when noisy target tokens attend to them.
Thus, DMD losses train how future noisy tokens read self-generated history, but not how that history is encoded into K/V memory.

\paragraph{SGF Pass 2: context-gradient reconstruction.}
SGF keeps Pass 1 and the reconstruction geometry unchanged, but removes the stop-gradient boundary on the reconstructed context K/V path.
The context latents $\tilde{X}_{\mathrm{ctx}}$ themselves remain stop-gradient inputs, so SGF does not optimize the sampled rollout trajectory.
Instead, the model re-encodes these fixed self-generated latents at $t_{\mathrm{ctx}}=0$, and the resulting K/V entries remain differentiable when future target tokens attend to them:
\begin{equation}
\begin{aligned}
\nabla_\theta \mathcal{L}_{\mathrm{DMD}}(\hat{X}_{\mathrm{tar}})
\supset
\frac{\partial \mathcal{L}_{\mathrm{DMD}}}{\partial \mathsf{KV}_{\mathrm{ctx}}^{\mathrm{rec}}}
\frac{\partial \mathsf{KV}_{\mathrm{ctx}}^{\mathrm{rec}}}{\partial \theta}.
\end{aligned}
\end{equation}
Therefore, future DMD losses supervise both target-side denoising and clean-context K/V writing.

The Pass-2 reconstruction is designed to recover the sampled exit computation from Pass 1.
With deterministic layers, matched positional indices, and the same causal reconstruction geometry, $\hat{X}_{\mathrm{tar}}$ is theoretically identical to the recorded predicted context latents $\tilde{X}_{\mathrm{ctx}}$ for the corresponding exit state.
In practice, small deviations may still appear due to floating-point and implementation-level effects.
Appendix~\ref{sec:two-pass-recovery} verifies this recovery fidelity empirically.

This two-pass design avoids the memory blow-up of a direct differentiable cache.
Pass 1 is serial but no-gradient; Pass 2 is gradient-enabled but fixed-window and parallel under $\mathcal{M}_{\mathrm{rec}}$.
SGF therefore recovers the missing memory-writing supervision without opening a recurrent autograd graph through the full self-rollout.

\subsection{Gradient Boundary}

SGF is a bounded reconstruction of the sampled exit computation, not full rollout BPTT.
The recorded context latents $\tilde{X}_{\mathrm{ctx}}$, noise samples, scheduler states, and Pass-1 serial cache trajectory are all stop-gradient.
Gradients flow only through the Pass-2 reconstruction: the clean-context forward, K/V projections, future-to-context attention, and target-side denoising computation.
This boundary recovers memory-writing supervision while keeping the training graph fixed-window and parallel.

\subsection{Streaming Context Policy}

For frame-wise streaming generation, both Self Forcing and SGF use the same sink-plus-FIFO context policy.
We keep a fixed sink prefix and a FIFO window of recent latents, so the evaluation isolates the effect of SGF rather than context selection.
For the Wan video VAE, we use four sink latents to preserve the temporal boundary prefix induced by its asymmetric grouping pattern.
Appendix~\ref{sec:context-policy-appendix} provides the boundary diagnostic and sink-number ablation.

\section{Experiments}
We evaluate whether \methodshort{} improves native long-video extrapolation without sacrificing short-horizon quality.
The appendix provides full 5s VBench tables, two-pass recovery fidelity, sink/context ablations, and additional qualitative comparisons (Appendix~\ref{sec:full-metric-tables}, \ref{sec:two-pass-recovery}, \ref{sec:context-policy-appendix}, and~\ref{sec:additional-qual}).

\subsection{Experimental Setup}
\label{sec:quantitative-analysis}

All models are trained with the same 5-second training window; 60s and 240s results therefore test native extrapolation beyond the training horizon.
Following prior long-video extrapolation practice~\citep{infinityrope}, we compare \methodshort{} with matched Self Forcing baselines at 5s, 60s, and 240s.
Each matched pair shares the same initialization, prompt set, random seed, sink/FIFO policy, sliding window, chunking strategy, and sampling configuration.
The only change is whether the sampled exit loss reads a frozen historical \kv{} cache, as in Self Forcing, or reconstructs the self-generated context with gradients, as in \methodshort{}.
For frame-wise generation, we use sink 4, total window 21, FIFO 16, and current chunk 1; for chunk-wise generation, we use sink 3, total window 12, FIFO 6, current chunk 3, and chunk size 3.

The 5s evaluation follows the standard VBench protocol and serves as a short-horizon sanity check.
Full 5s VBench results are reported in Appendix~\ref{sec:full-metric-tables}, where \methodshort{} and Self Forcing are broadly comparable across frame-wise and chunk-wise settings.
We therefore focus the main paper on 60s and 240s extrapolation, where memory-writing errors have time to accumulate.

The 60s setting uses VBench-Long prompts, and the 240s setting uses 128 randomly sampled MovieGen prompts~\citep{moviegen}.
For both long horizons, we report the official VBench-Long metrics: aesthetic quality, background consistency, dynamic degree, imaging quality, motion smoothness, subject consistency, and flickering.
Higher is better after VBench orientation.
GSB human-preference results are reported separately as paired comparisons.



\subsection{Long Video Generation (60s and 240s)}

\paragraph{Quantitative results.}
Tabs.~\ref{tab:framewise-long} and~\ref{tab:chunkwise-long} report long-horizon automatic metrics for frame-wise and chunk-wise autoregressive generation.
All comparisons are controlled within each generation setting and should be read pairwise by initialization.

Across both generation granularities and both long horizons, \methodshort{} improves over matched Self Forcing baselines on most quality and consistency metrics.
The gains are especially consistent for aesthetic quality, background consistency, imaging quality, motion smoothness, subject consistency, and flickering.
These metrics directly reflect the failure mode targeted by \methodshort{}: whether self-generated latents are written into memory representations that remain useful throughout long extrapolation.

Dynamic degree is the main exception, where Self Forcing can obtain a higher score.
This does not necessarily indicate better motion quality: Appendix~\ref{sec:additional-qual} shows that long Self Forcing rollouts often contain scene jumps, broken camera geometry, and object deformation, which create large but incoherent apparent motion and can inflate dynamic degree.
By contrast, \methodshort{} maintains more stable image quality and more plausible camera evolution, so its lower dynamic-degree score in these cases is not evidence of worse generation quality.

\begin{figure}[t]
\centering
\includegraphics[width=\linewidth]{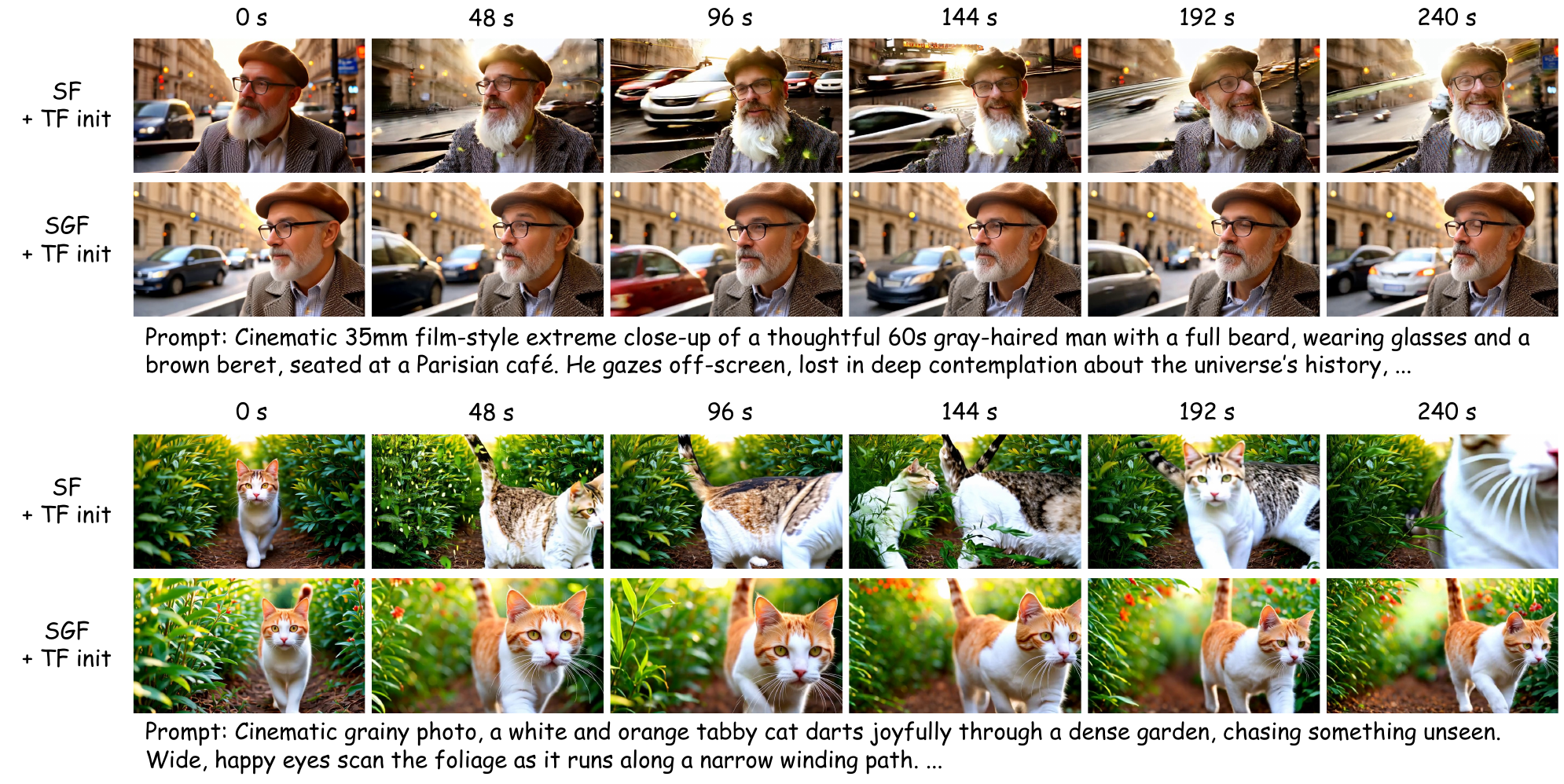}
\vspace{-3mm}
\caption{
\textbf{Frame-wise 240-second comparison under TF initialization.}
Both rows use matched prompts, random seeds, initialization, horizon, and inference geometry.
}
\vspace{-4mm}
\label{fig:framewise-tf-240-main}
\end{figure}

\paragraph{Qualitative results.}
Fig.~\ref{fig:framewise-tf-240-main} shows a representative 240-second frame-wise comparison under TF initialization.
Both methods use the same prompts, seeds, horizon, and inference geometry, so the visual differences reflect the training objective rather than sampling variation.
Self Forcing remains locally plausible early in the rollout, but gradually accumulates identity drift, crop drift, and layout changes, indicating that its self-generated K/V memory becomes less reliable for later latent generation.
In contrast, \methodshort{} uses future losses to supervise the clean-context K/V representations of self-generated latents, better preserving subject identity, camera relation, and textured background over the full rollout.
Additional frame-wise and chunk-wise comparisons under other initializations are provided in Appendix~\ref{sec:additional-qual}.

\begin{table}[t]
\centering
\caption{Frame-wise 60-second and 240-second long-horizon metrics. The 60s setting uses VBench-Long prompts; the 240s setting uses MovieGen-128 prompts.}
\label{tab:framewise-long}
\footnotesize
\setlength{\tabcolsep}{2.5pt}
\begin{tabularx}{\linewidth}{@{}YY|YY|YY|YY@{}}
\toprule
Horizon & Metric$\uparrow$ & \multicolumn{2}{c|}{Causal ODE init} & \multicolumn{2}{c|}{Causal CD init} & \multicolumn{2}{c}{TF init} \\
\cmidrule(lr){3-4}\cmidrule(lr){5-6}\cmidrule(lr){7-8}
& & \sfcell & SGF & \sfcell & SGF & \sfcell & SGF \\
\midrule
\multirow{7}{*}{60s}
& Aesthetics & 0.543 & \textbf{0.606} & 0.608 & \textbf{0.630} & 0.650 & \textbf{0.653} \\
& Background & 0.947 & \textbf{0.966} & 0.969 & \textbf{0.973} & 0.970 & \textbf{0.974} \\
& Dynamics & \textbf{0.867} & 0.566 & \textbf{0.489} & 0.267 & 0.647 & \textbf{0.730} \\
& Imaging & 0.657 & \textbf{0.715} & 0.684 & \textbf{0.708} & 0.704 & \textbf{0.714} \\
& Motion & 0.975 & \textbf{0.986} & 0.984 & \textbf{0.988} & \textbf{0.985} & 0.982 \\
& Subject & 0.928 & \textbf{0.971} & 0.974 & \textbf{0.983} & 0.976 & \textbf{0.983} \\
& Flickering & 0.971 & \textbf{0.987} & 0.992 & \textbf{0.992} & 0.988 & \textbf{0.991} \\
\midrule
\multirow{7}{*}{240s}
& Aesthetics & 0.510 & \textbf{0.543} & 0.533 & \textbf{0.584} & 0.614 & \textbf{0.619} \\
& Background & 0.943 & \textbf{0.966} & 0.964 & \textbf{0.971} & 0.965 & \textbf{0.968} \\
& Dynamics & \textbf{0.840} & 0.556 & \textbf{0.599} & 0.385 & \textbf{0.669} & 0.648 \\
& Imaging & 0.648 & \textbf{0.698} & 0.632 & \textbf{0.690} & 0.701 & \textbf{0.718} \\
& Motion & 0.971 & \textbf{0.985} & 0.981 & \textbf{0.986} & 0.980 & \textbf{0.982} \\
& Subject & 0.936 & \textbf{0.969} & 0.967 & \textbf{0.976} & 0.965 & \textbf{0.972} \\
& Flickering & 0.937 & \textbf{0.972} & 0.971 & \textbf{0.977} & 0.967 & \textbf{0.968} \\
\bottomrule
\end{tabularx}
\vspace{-4mm}
\end{table}

\begin{table}[t]
\centering
\caption{Chunk-wise 60-second and 240-second long-horizon metrics. 60s uses VBench-Long prompts; 240s uses MovieGen-128 prompts.}
\label{tab:chunkwise-long}
\footnotesize
\setlength{\tabcolsep}{1pt}
\begin{tabularx}{0.96\linewidth}{@{}YY|Y|Y|YY|YY@{}}
\toprule
Horizon & Metric$\uparrow$ & \shortstack{Bidirectional\\ODE init} & \shortstack{Causal\\ODE init} & \multicolumn{2}{c|}{Causal CD init} & \multicolumn{2}{c}{TF init} \\
\cmidrule(lr){3-3}\cmidrule(lr){4-4}\cmidrule(lr){5-6}\cmidrule(lr){7-8}
& & \sfcell & \sfcell & \sfcell & SGF & \sfcell & SGF \\
\midrule
\multirow{7}{*}{60s}
& Aesthetics & 0.605 & 0.599 & 0.603 & \textbf{0.605} & 0.582 & \textbf{0.654} \\
& Background & 0.966 & 0.952 & 0.957 & \textbf{0.958} & 0.947 & \textbf{0.971} \\
& Dynamics & 0.613 & 0.935 & 0.832 & \textbf{0.913} & \textbf{0.909} & 0.634 \\
& Imaging & 0.692 & 0.700 & 0.677 & \textbf{0.681} & 0.686 & \textbf{0.715} \\
& Motion & 0.983 & 0.961 & 0.974 & \textbf{0.976} & 0.968 & \textbf{0.985} \\
& Subject & 0.974 & 0.948 & \textbf{0.960} & \textbf{0.960} & 0.951 & \textbf{0.982} \\
& Flickering & 0.991 & 0.993 & 0.988 & \textbf{0.989} & 0.989 & \textbf{0.990} \\
\midrule
\multirow{7}{*}{240s}
& Aesthetics & 0.559 & 0.564 & 0.569 & \textbf{0.578} & 0.557 & \textbf{0.629} \\
& Background & 0.962 & 0.941 & 0.948 & \textbf{0.950} & 0.944 & \textbf{0.970} \\
& Dynamics & 0.622 & 0.916 & 0.823 & \textbf{0.923} & \textbf{0.943} & 0.566 \\
& Imaging & 0.668 & 0.679 & \textbf{0.671} & 0.667 & 0.688 & \textbf{0.712} \\
& Motion & 0.977 & 0.957 & 0.967 & \textbf{0.970} & 0.967 & \textbf{0.985} \\
& Subject & 0.962 & 0.939 & 0.939 & \textbf{0.942} & 0.946 & \textbf{0.975} \\
& Flickering & 0.959 & 0.915 & 0.941 & \textbf{0.943} & 0.934 & \textbf{0.970} \\
\bottomrule
\end{tabularx}
\vspace{-4mm}
\end{table}

\subsection{User Study}

Automatic metrics do not directly capture human-perceived long-horizon consistency, so we conduct a blind GSB preference study with more than 1,900 paired judgments across 10 matched comparisons.
Each pair compares \methodshort{} against the matched Self Forcing baseline under the same initialization, prompt set, horizon, and inference geometry.
We report $\mathrm{GSB}=(G-B)/(G+S+B)\times100\%$, where $G$ favors \methodshort{}, $B$ favors Self Forcing, and $S$ indicates no clear preference; positive scores therefore indicate preference for \methodshort{}.
As shown in Tab.~\ref{tab:long-gsb}, all scores are positive, indicating that raters consistently prefer \methodshort{} over Self Forcing in long-horizon generation.

\begin{table}[t]
\centering
\caption{Long-horizon GSB preference scores for \methodshort{} vs. matched Self Forcing baselines in frame-wise and chunk-wise generation. Each cell reports $(G-B)/(G+S+B)\times100\%$ under the same initialization, prompt set, rollout horizon, and inference geometry. Positive values indicate preference for \methodshort{} over Self Forcing.}
\label{tab:long-gsb}
\footnotesize
\setlength{\tabcolsep}{4pt}
\begin{tabularx}{\linewidth}{@{}YY|Y|Y|Y@{}}
\toprule
Setting & Horizon & Causal ODE init & Causal CD init & TF init \\
\midrule
Frame-wise & 60s &  29.6\% &  36.8\% & 45.1\% \\
Frame-wise & 240s &  38.6\% & 35.9\% & 48.7\% \\
Chunk-wise & 60s & -- & 32.9\% & 37.5\% \\
Chunk-wise & 240s & -- & 34.3\% & 44.9\% \\
\bottomrule
\end{tabularx}
\vspace{-4mm}
\end{table}

\subsection{Qualitative Results}
\label{sec:qualitative-analysis}

We summarize the long-horizon qualitative comparisons provided in Appendix~\ref{sec:additional-qual}.
Each strip compares \methodshort{} and Self Forcing under the same prompt, random seed, initialization, horizon, and inference geometry, so the visual differences reflect the training objective rather than sampling variation.
These strips visualize the temporal failure mode targeted by \methodshort{}: a generated history can remain locally plausible while becoming progressively less useful as memory for later latents.
The TF-initialized frame-wise 60s and 240s comparisons in Figs.~\ref{fig:framewise-tf-60-main} and~\ref{fig:framewise-tf-240-main} provide the clearest examples, with additional frame-wise and chunk-wise cases covering other initializations.

Across the appendix examples, Self Forcing often accumulates long-horizon drift, including scene jumps, broken camera geometry, object deformation, identity drift, crop drift, and background/layout replacement.
These failures can produce large but incoherent apparent motion, explaining why dynamic degree may increase despite worse perceptual quality.
By contrast, \methodshort{} better preserves subject identity, camera relation, scene layout, and temporal stability, matching the gains in subject consistency, background consistency, flickering, and motion smoothness.

\subsection{Training Feasibility}

We analyze whether recovering context gradients makes training prohibitively expensive.
Tab.~\ref{tab:analysis-memory-feasibility} compares frozen-cache Self Forcing, \methodshort{}, and a direct differentiable-cache variant.
Directly enabling gradients through the historical \kv{} cache runs out of memory, because each cached entry retains the serial cache-formation graph that produced it.
\methodshort{} avoids this recurrent graph: Pass 1 is a no-gradient rollout, and gradients are opened only in the bounded Pass-2 reconstruction.
In our implementation, Pass 2 uses FlexAttention with a compiled static block-sparse causal mask, making the reconstructed future-to-context attention memory efficient.

The measured overhead is modest.
Compared with frozen-cache Self Forcing, \methodshort{} increases peak memory from 79.01GB to 87.01GB, while stable memory decreases from 79.01GB to 63.73GB.
Thus, activating the context-gradient path does not cause the recurrent memory growth of direct differentiable-cache training.
Runtime is similarly close: wall-clock time increases from 10.39s to 11.71s per five training steps.
This is because, for exit step $n$, Self Forcing performs $n-1$ no-gradient generator forwards and one gradient-enabled forward/backward at the exit step, whereas \methodshort{} performs $n$ no-gradient forwards in Pass 1 and one gradient-enabled forward/backward in Pass 2.
Since fake-score and generator updates follow a 5:1 schedule, the extra Pass-2 work affects only the generator update within each five-step cycle.

\begin{table}[t]
\centering
\caption{
\textbf{Training feasibility of context-gradient recovery.}
Direct differentiable-cache training keeps the serial cache-formation graph and runs out of memory.
\methodshort{} uses bounded parallel context-gradient reconstruction at the sampled exit step, restoring context K/V gradients with modest peak-memory and runtime overhead.
}
\label{tab:analysis-memory-feasibility}
\small
\setlength{\tabcolsep}{4pt}
\begin{tabular*}{\linewidth}{@{\extracolsep{\fill}}lcccc}
\toprule
Training variant & Peak memory & Stable memory & Time / 5 steps & Outcome \\
\midrule
Self Forcing, frozen historical \kv{} cache & 79.01GB & 79.01GB & 10.39s & trains \\
\methodshort{}, serial Pass 1 + parallel Pass 2 & 87.01GB & 63.73GB & 11.71s & trains \\
Self Forcing, differentiable historical \kv{} cache & OOM & OOM & -- & OOM \\
\bottomrule
\end{tabular*}
\end{table}
\section{Conclusion}

We presented \method{} (\methodshort{}), a two-pass training strategy that closes the historical context-gradient gap in frozen-cache Self Forcing. By keeping the serial self-rollout no-gradient and reconstructing the sampled exit computation in parallel, \methodshort{} lets future losses supervise how self-generated histories are written into K/V memory without full rollout backpropagation. Across frame-wise and chunk-wise generation, multiple initializations, and 5s, 60s, and 240s horizons, \methodshort{} preserves short-horizon quality while improving long-video identity, layout consistency, and temporal stability. These gains are confirmed by qualitative comparisons and human preference, and future work can combine \methodshort{} with stronger initialization, long-context tuning, retrieval, and cache-compression techniques.

\bibliography{references}
\bibliographystyle{iclr2026_conference}

\appendix
\section{Additional Experimental Details}

\paragraph{Benchmark protocol.}
We evaluate matched Self Forcing and \methodshort{} pairs at 5s, 60s, and 240s. The 5-second setting follows the standard VBench protocol and reports all 16 quality and semantic dimensions. The 60-second setting uses the VBench-Long protocol, while the 240-second setting uses 128 randomly sampled MovieGen prompts with the same VBench-Long quality metrics. For long-horizon evaluation, we omit text-alignment metrics and focus on visual persistence under autoregressive extrapolation.

\paragraph{Self Forcing checkpoint provenance.}
The frame-wise and chunk-wise causal-ODE Self Forcing baselines use the released Causal Forcing checkpoints~\citep{causalforcing}. The chunk-wise bidirectional-ODE Self Forcing baseline uses the released Self Forcing checkpoint~\citep{selfforcing}. All other Self Forcing checkpoints in our tables are reproduced by us under the same training setting as the corresponding \methodshort{} checkpoints; evaluation prompts, seeds, sampling configuration, and inference context geometry are matched within each comparison.

\subsection{Frame-wise configuration}

For frame-wise training, we use the full 5-second training window without sliding-window eviction. Accordingly, the Pass-2 reconstruction uses a standard teacher-forcing-style causal mask over the recorded frame sequence, rather than a sink-plus-FIFO sliding-window mask. This mask matches the full-context causal relation used by the frame-wise Pass-1 training rollout.

At inference and long-horizon evaluation time, frame-wise generation is run in streaming mode with sink 4, FIFO 16, and current chunk 1, giving a total context window of 21 latent frames. This streaming policy is shared by Self Forcing and \methodshort{}; within each matched pair, the prompt set, random seed, sampling configuration, initialization, and inference context policy are held fixed. The only intended difference is the gradient boundary of the sampled exit loss: Self Forcing consumes the historical \kv{} cache as frozen rollout state, whereas \methodshort{} reconstructs the self-generated context with gradients through the clean-context K/V path.

The corresponding 5-second results are reported in Tab.~\ref{tab:framewise-5s}, and the 60-second and 240-second results are reported in Tab.~\ref{tab:framewise-long}.

\subsection{Chunk-wise configuration}

For chunk-wise training, we use a sliding-window context during the Pass-1 rollout. The window contains sink 3, FIFO 6, and current chunk 3, with chunk size 3. The Pass-2 reconstruction mask is built to match this Pass-1 sliding-window cache relation, so the context-gradient recovery follows the same attention geometry used during the sampled rollout.

The same chunk-wise context policy is used for inference and long-horizon evaluation. We include released bidirectional-ODE Self Forcing and causal-ODE Self Forcing checkpoints as reference baselines. The controlled \methodshort{} comparisons are the matched causal-CD and TF pairs, where Self Forcing and \methodshort{} share the same initialization and inference geometry.

This setting tests whether the context-gradient signal remains useful when memory is updated at a coarser temporal granularity. The 5-second chunk-wise results are reported in Tab.~\ref{tab:app-chunkwise-5s}, and the 60-second and 240-second results are reported in Tab.~\ref{tab:chunkwise-long}. Human preference scores are reported separately in Tab.~\ref{tab:long-gsb}.
\section{Full Metric Tables}
\label{sec:full-metric-tables}

This appendix reports the complete 5-second VBench results. Metrics are rows and model variants are columns, and boldface marks the better value within each matched Self Forcing--SGF pair. These short-horizon results are used as a sanity check: SGF is designed to improve long autoregressive extrapolation, so it should not degrade standard 5-second video quality. The long-horizon 60s and 240s results are reported in the main paper.

\subsection{Frame-wise 5-second VBench}

The frame-wise 5-second setting evaluates whether adding the context-gradient reconstruction objective preserves short-video generation quality. Frame-wise training uses the full 5-second training window without sliding-window eviction, and the Pass-2 reconstruction uses a standard teacher-forcing-style causal mask. At evaluation time, we use the same streaming policy as the long-horizon frame-wise experiments: sink 4, FIFO 16, and current chunk 1, giving a total context window of 21 latent frames. For 5-second videos, this window covers the full generated latent sequence, so the table mainly measures short-horizon quality rather than long-range extrapolation.

\begin{table}[t]
\centering
\caption{Frame-wise 5-second VBench metrics. Evaluation uses sink 4, FIFO 16, and current chunk 1, for a total context window of 21 latent frames. Paired columns isolate the SGF gradient-boundary change under the same initialization.}
\label{tab:framewise-5s}
\footnotesize
\setlength{\tabcolsep}{3pt}
\begin{tabularx}{\linewidth}{@{}YYY|YY|YY@{}}
\toprule
Metric & \multicolumn{2}{c|}{Causal ODE init} & \multicolumn{2}{c|}{Causal CD init} & \multicolumn{2}{c}{TF init} \\
\cmidrule(lr){2-3}\cmidrule(lr){4-5}\cmidrule(lr){6-7}
& \sfcell & SGF & \sfcell & SGF & \sfcell & SGF \\
\midrule
Aesthetics & 0.651 & \textbf{0.665} & 0.663 & \textbf{0.672} & 0.667 & \textbf{0.671} \\
Background & 0.928 & \textbf{0.956} & 0.963 & \textbf{0.968} & \textbf{0.959} & \textbf{0.959} \\
Dynamics & \textbf{0.989} & 0.703 & \textbf{0.616} & 0.375 & 0.633 & \textbf{0.653} \\
Imaging & 0.694 & \textbf{0.713} & 0.711 & \textbf{0.714} & 0.698 & \textbf{0.713} \\
Motion & 0.972 & \textbf{0.985} & 0.983 & \textbf{0.989} & \textbf{0.986} & 0.983 \\
Subject & 0.911 & \textbf{0.963} & 0.953 & \textbf{0.976} & 0.961 & \textbf{0.968} \\
Flickering & 0.947 & \textbf{0.984} & \textbf{0.993} & \textbf{0.993} & 0.988 & \textbf{0.990} \\
Object & 0.938 & \textbf{0.951} & 0.955 & \textbf{0.960} & 0.941 & \textbf{0.950} \\
Multiple & 0.789 & \textbf{0.865} & 0.861 & \textbf{0.883} & 0.849 & \textbf{0.855} \\
Action & \textbf{0.968} & 0.962 & \textbf{0.954} & 0.950 & \textbf{0.962} & 0.958 \\
Color & 0.842 & \textbf{0.863} & \textbf{0.885} & 0.883 & 0.847 & \textbf{0.882} \\
Spatial & 0.732 & \textbf{0.778} & 0.771 & \textbf{0.780} & \textbf{0.781} & 0.739 \\
Scene & 0.563 & \textbf{0.567} & 0.574 & \textbf{0.584} & 0.538 & \textbf{0.547} \\
Appearance & \textbf{0.206} & 0.204 & 0.199 & \textbf{0.204} & \textbf{0.204} & 0.203 \\
Temporal & 0.246 & \textbf{0.250} & \textbf{0.242} & \textbf{0.242} & \textbf{0.239} & \textbf{0.239} \\
Consistency & \textbf{0.264} & 0.262 & \textbf{0.263} & \textbf{0.263} & 0.262 & \textbf{0.264} \\
\bottomrule
\end{tabularx}
\end{table}

\subsection{Chunk-wise 5-second VBench}

The chunk-wise 5-second setting checks whether SGF remains comparable to Self Forcing when memory is updated at a coarser temporal granularity. Unlike frame-wise training, chunk-wise training already uses a sliding-window context in Pass 1, with sink 3, FIFO 6, current chunk 3, and chunk size 3. The Pass-2 reconstruction mask matches this sliding-window cache relation, so the recovered context-gradient path follows the same attention geometry as the sampled rollout. The released bidirectional-ODE and causal-ODE Self Forcing rows are included as reference baselines, while the controlled SGF comparisons are the matched causal-CD and TF pairs.

\begin{table}[t]
\centering
\caption{Chunk-wise 5-second VBench metrics. Training and evaluation use sink 3, FIFO 6, current chunk 3, and chunk size 3. The controlled SGF comparisons are the matched causal-CD and TF pairs; ODE-initialized Self Forcing rows are reference baselines.}
\label{tab:app-chunkwise-5s}
\footnotesize
\setlength{\tabcolsep}{3pt}
\begin{tabularx}{\linewidth}{@{}YY|Y|YY|YY@{}}
\toprule
Metric & \shortstack{Bidirectional\\ODE init} & \shortstack{Causal\\ODE init} & \multicolumn{2}{c|}{Causal CD init} & \multicolumn{2}{c}{TF init} \\
\cmidrule(lr){2-2}\cmidrule(lr){3-3}\cmidrule(lr){4-5}\cmidrule(lr){6-7}
& \sfcell & \sfcell & \sfcell & SGF & \sfcell & SGF \\
\midrule
Aesthetics & 0.659 & 0.660 & 0.661 & \textbf{0.663} & 0.663 & \textbf{0.678} \\
Background & 0.961 & 0.959 & 0.945 & \textbf{0.958} & 0.948 & \textbf{0.964} \\
Dynamics & 0.647 & 0.836 & 0.753 & \textbf{0.875} & \textbf{0.908} & 0.692 \\
Imaging & 0.694 & 0.705 & \textbf{0.699} & \textbf{0.699} & 0.695 & \textbf{0.706} \\
Motion & 0.984 & 0.974 & \textbf{0.978} & 0.975 & 0.976 & \textbf{0.985} \\
Subject & 0.955 & 0.955 & 0.946 & \textbf{0.956} & 0.938 & \textbf{0.970} \\
Flickering & 0.991 & 0.982 & 0.979 & \textbf{0.981} & 0.978 & \textbf{0.988} \\
Object & 0.952 & 0.958 & 0.951 & \textbf{0.958} & \textbf{0.958} & \textbf{0.958} \\
Multiple & 0.863 & 0.866 & \textbf{0.844} & 0.816 & 0.852 & \textbf{0.861} \\
Action & 0.968 & 0.956 & \textbf{0.962} & 0.952 & 0.950 & \textbf{0.964} \\
Color & 0.880 & 0.879 & 0.880 & \textbf{0.890} & 0.872 & \textbf{0.875} \\
Spatial & 0.811 & 0.791 & 0.741 & \textbf{0.752} & 0.779 & \textbf{0.802} \\
Scene & 0.574 & 0.558 & \textbf{0.560} & 0.557 & 0.542 & \textbf{0.551} \\
Appearance & 0.203 & 0.205 & \textbf{0.200} & 0.199 & 0.202 & \textbf{0.203} \\
Temporal & 0.244 & 0.247 & \textbf{0.245} & 0.244 & 0.241 & \textbf{0.242} \\
Consistency & 0.268 & 0.267 & \textbf{0.265} & \textbf{0.265} & \textbf{0.266} & 0.265 \\
\bottomrule
\end{tabularx}
\end{table}

\section{Direct Cache-Gradient Feasibility}
\label{sec:memory-feasibility}

This appendix expands on the training-feasibility discussion in the main paper. The measured comparison is reported in Tab.~\ref{tab:analysis-memory-feasibility}. Here we separate three possible ways to expose gradients through autoregressive history: frozen-cache Self Forcing, direct differentiable-cache training without denoising-trajectory gradients, and full rollout BPTT. This distinction clarifies why SGF uses bounded parallel reconstruction.

\paragraph{Frozen-cache Self Forcing.}
In frozen-cache Self Forcing, the sampled exit-step prediction is trained with gradients, but the historical cache update is detached. Non-exit denoising steps are executed without gradient tracking, and after each generated block is predicted, the model is called again at the context timestep to update the persistent \kv{} cache under no-gradient execution. Thus future losses can train the target-side denoising computation that reads historical cache entries, but they cannot backpropagate into the $t_{\mathrm{ctx}}=0$ forward computation that produced those historical \kv{} entries.

For batch size $B$, retained historical blocks $T$, tokens per block $L$, width $D$, and $H$ transformer layers, the raw detached cache storage scales as
\begin{equation}
M_{\mathrm{cache}}
\approx
H \cdot 2 \cdot B \cdot T L D \cdot \mathrm{bytes},
\label{eq:appendix-cache-data}
\end{equation}
where the factor 2 accounts for keys and values. This term is only the tensor footprint of the retained cache, not the activation graph that would be needed to differentiate through its formation.

\paragraph{Direct differentiable cache without denoising-trajectory gradients.}
A direct way to recover the missing history-formation gradient is to keep the persistent historical cache differentiable, while still treating the generated latents themselves as stop-gradient inputs. In this variant,
\begin{equation}
\mathsf{KV}_i^0(\theta)
=
\mathcal{C}_\theta\!\left(
\operatorname{sg}(\tilde{x}_i), t_{\mathrm{ctx}}; \mathsf{KV}_{<i}^0
\right),
\qquad
t_{\mathrm{ctx}}=0,
\end{equation}
where $\mathcal{C}_\theta$ denotes the same causal model forward that produces the historical K/V entries. Although $\tilde{x}_i$ is detached, the cache formation itself remains serial: the K/V entry for block $i$ is computed while reading earlier historical cache entries. If those earlier entries are also differentiable, the autograd graph becomes recurrent across the rollout.

The memory cost is therefore larger than the raw cache storage:
\begin{equation}
M_{\mathrm{direct}}
\gtrsim
M_{\mathrm{cache}}
+
\sum_{i=1}^{T}
M_{\mathrm{KV\ formation}}(i)
+
M_{\mathrm{saved\ attention}},
\label{eq:appendix-direct-memory}
\end{equation}
where $M_{\mathrm{KV\ formation}}(i)$ denotes the saved activations of the $t_{\mathrm{ctx}}=0$ forward computation that forms the historical K/V entry for block $i$, including its dependence on earlier cache state. This expression is a scaling argument rather than an allocator-exact formula. Its purpose is to show that even the minimal direct-cache alternative, with generated latents detached, opens a serial history-formation graph.

\paragraph{Full rollout BPTT.}
A still stronger alternative is full rollout BPTT, where the generated latents are not detached. Then the historical K/V entry for block $i$ depends not only on the $t_{\mathrm{ctx}}=0$ K/V-forming forward, but also on the denoising trajectory that produced $\tilde{x}_i$. The memory cost further includes the saved activations of the denoising steps:
\begin{equation}
M_{\mathrm{full\text{-}BPTT}}
\gtrsim
M_{\mathrm{direct}}
+
\sum_{i=1}^{T}
\sum_{s=1}^{K_i}
M_{\mathrm{denoise}}(i,s),
\label{eq:appendix-full-bptt-memory}
\end{equation}
where $K_i$ is the number of denoising steps used for block $i$, and $M_{\mathrm{denoise}}(i,s)$ denotes the activation footprint of the generator forward at denoising step $s$. Full rollout BPTT therefore grows across both rollout length and denoising depth, making it strictly more demanding than the direct differentiable-cache variant above.

\paragraph{SGF bounded reconstruction.}
SGF avoids retaining either serial graph. Pass 1 performs the true autoregressive rollout under no-gradient execution and records the self-generated context latents $\tilde{X}_{\mathrm{ctx}}$ together with the sampled noisy exit states $Z^{t^\star}$. Pass 2 discards the persistent rollout cache and performs one bounded parallel reconstruction:
\begin{equation}
M_{\methodshort{}}
\approx
M_{\mathrm{pass1\ cache\ data}}
+
M_{\mathrm{records}}(\tilde{X}_{\mathrm{ctx}}, Z^{t^\star})
+
M_{\mathrm{parallel\ window}}(N),
\label{eq:appendix-sgf-memory}
\end{equation}
where $N$ is the fixed reconstruction-window length. The recovered gradient passes through the context-side forward at $t_{\mathrm{ctx}}=0$, the K/V projections, and the future-to-context attention relation in Pass 2. It does not backpropagate through the denoising trajectory that produced the recorded latents, nor through the serial persistent-cache updates from Pass 1.

\paragraph{Interpretation.}
This gives the relevant scaling order:
\begin{equation}
\begin{aligned}
M_{\methodshort{}}
<
M_{\mathrm{direct}}
<
M_{\mathrm{full\text{-}BPTT}},
\end{aligned}
\end{equation}
where the middle term refers to direct differentiable-cache training with stop-gradient generated latents, not the frozen-cache Self Forcing baseline. Frozen-cache Self Forcing can be cheaper because it leaves the historical K/V formation path detached, but that is exactly the missing gradient SGF is designed to recover. The measured results in Tab.~\ref{tab:analysis-memory-feasibility} are consistent with this picture: direct differentiable-cache training runs out of memory in our setting, while SGF restores the context-gradient path through a bounded exit-step replay.
\section{Two-Pass Recovery Fidelity}
\label{sec:two-pass-recovery}

SGF relies on Pass 2 being a faithful replay of the sampled exit computation from Pass 1. This appendix verifies that assumption at the forward level. The goal is not to claim equivalence to full rollout BPTT, but to check that the parallel reconstruction reproduces the same local causal relation as the serial no-gradient rollout, up to the numerical differences expected from mixed-precision execution.

\paragraph{Protocol.}
We evaluate recovery fidelity on 24 prompts and four exit steps, $t^\star \in \{1000,750,500,250\}$, giving 96 prompt/exit comparisons. For each comparison, Pass 1 performs the serial self-rollout under no-gradient execution and records the sampled noisy input together with the corresponding exit-step prediction. Pass 2 then performs one parallel reconstruction using the recorded noisy states and detached self-generated context under the matched teacher-forcing attention mask. We compare the Pass-1 and Pass-2 latent predictions before VAE decoding.

\paragraph{Metrics.}
Let $x^{(1)}$ denote the Pass-1 exit prediction and $x^{(2)}$ denote the corresponding Pass-2 reconstruction. We report mean squared error, root mean squared error, mean absolute error, maximum absolute error, relative $\ell_2$ error, and cosine similarity:
\begin{align}
\mathrm{RMSE} &= \sqrt{\operatorname{mean}\left((x^{(2)} - x^{(1)})^2\right)}, \\
\mathrm{RelL2} &= \frac{\lVert x^{(2)} - x^{(1)} \rVert_2}{\max(\lVert x^{(1)} \rVert_2, 10^{-12})}, \\
\mathrm{Cosine} &= \frac{\langle x^{(1)}, x^{(2)} \rangle}{\lVert x^{(1)} \rVert_2 \lVert x^{(2)} \rVert_2}.
\end{align}
To relate the recovery error to mixed-precision numerical scale, we also report $\mathrm{RelL2}/\epsilon_{\mathrm{bf16}}$, where $\epsilon_{\mathrm{bf16}}=2^{-7}$ is the relative precision scale of bf16. Since both serial cache execution and parallel reconstruction pass through many bf16 transformer operations, accumulated roundoff can naturally be a small multiple of this reference scale.

\begin{table}[t]
\centering
\caption{Pass-1 versus Pass-2 latent recovery fidelity. Metrics are averaged over 24 prompts for each exit step. Overall averages are computed over all 96 prompt/exit comparisons. We also report the relative $\ell_2$ error normalized by the bf16 relative precision $\epsilon_{\mathrm{bf16}}=2^{-7}$.}
\label{tab:recovery-error-appendix}
\resizebox{\linewidth}{!}{%
\begin{tabular}{lcccccccc}
\toprule
Exit step & \# comparisons & MSE & RMSE & Mean abs. & Max abs. & Rel. L2 & Rel. L2 / $\epsilon_{\mathrm{bf16}}$ & Cosine \\
\midrule
1000 & 24 & 3.123e-4 & 0.01745 & 0.01093 & 0.58396 & 0.02133 & 2.73 & 0.999766 \\
750  & 24 & 2.148e-4 & 0.01449 & 0.00884 & 0.58189 & 0.01566 & 2.00 & 0.999874 \\
500  & 24 & 1.141e-4 & 0.01064 & 0.00693 & 0.49110 & 0.01125 & 1.44 & 0.999936 \\
250  & 24 & 6.041e-5 & 0.00774 & 0.00535 & 0.29028 & 0.00812 & 1.04 & 0.999967 \\
\midrule
Overall & 96 & 1.754e-4 & 0.01258 & 0.00801 & 0.48681 & 0.01409 & 1.80 & 0.999886 \\
\bottomrule
\end{tabular}}
\end{table}

\paragraph{Results.}
Tab.~\ref{tab:recovery-error-appendix} shows that Pass 2 closely reproduces the Pass-1 exit predictions across the denoising schedule. The overall relative $\ell_2$ error is 1.41\%, and the average cosine similarity is 0.999886. The relative $\ell_2$ error is also close to the bf16 numerical scale: overall it is 1.80 times $\epsilon_{\mathrm{bf16}}$, and the ratio decreases from 2.73 at the noisiest exit step to 1.04 at exit step 250. Because the two paths execute many bf16 transformer operations with different serial and parallel computation orders, such small multiples of the bf16 reference scale are consistent with accumulated floating-point roundoff rather than a substantive mismatch in the recovered computation.

These results support the use of Pass 2 as a bounded local surrogate for the sampled exit computation. The diagnostic verifies forward recovery of the local attention computation up to expected mixed-precision numerical error; it does not imply that SGF recovers the exact gradient of the full serial rollout.

\begin{figure}[t]
\centering
\includegraphics[width=\linewidth]{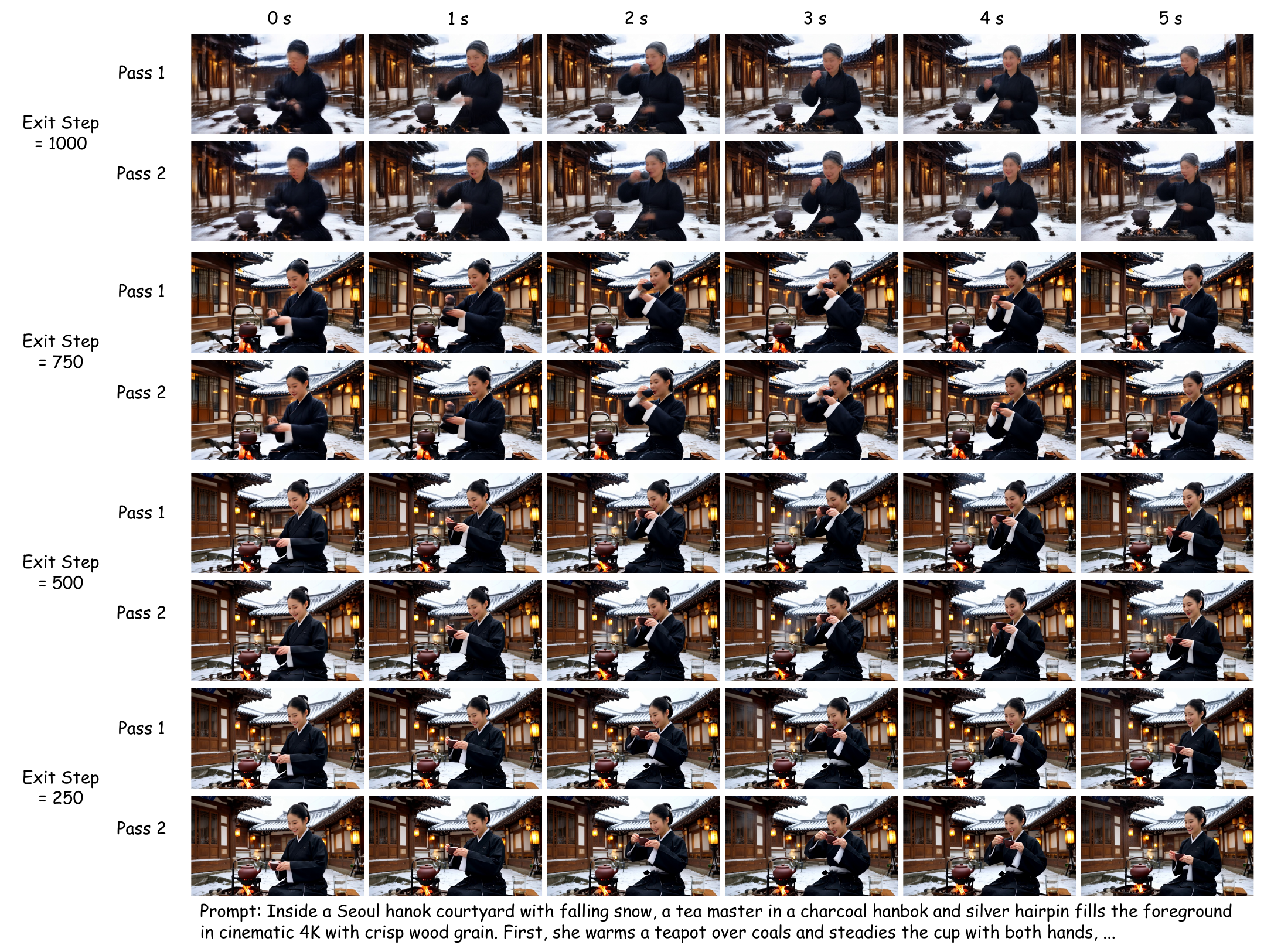}
\caption{\textbf{Decoded Pass-1 and Pass-2 recovery comparison.} Rows show paired decoded outputs from Pass 1 and Pass 2 at exit steps 1000, 750, 500, and 250. The decoded pairs are visually nearly indistinguishable, matching the latent-space recovery results in Tab.~\ref{tab:recovery-error-appendix}.}
\label{fig:twopass-appendix}
\end{figure}
\section{VAE Boundary and Sink Choice}
\label{sec:context-policy-appendix}

Frame-wise streaming inference uses a sink-plus-FIFO context policy. This policy is shared by Self Forcing and \methodshort{}, so it is not an independent SGF contribution. We include this appendix to justify the sink size used in the frame-wise experiments.

\paragraph{Wan VAE start-boundary diagnostic.}
The Wan video VAE encodes time with an asymmetric grouping pattern: the stream starts with a 1-frame boundary group, followed by 4-frame groups. As a result, the first few latent positions of a freshly encoded clip are not necessarily equivalent to steady in-stream latents. To measure this effect, we use 10 videos with 81 pixel-space frames, corresponding to 21 VAE latent frames. For each video, we encode the full prefix and take the 21st latent, i.e., zero-based index 20, as the reference.

We then re-encode local windows ending at the same target frame group and compare the last local latent to the full-prefix reference. The fresh 4-frame control uses only frames [77, 81), but because the VAE starts every fresh stream with a 1-frame boundary group, this produces a start-boundary latent rather than a normal in-stream 4-frame latent. The anchor-window setting includes one boundary anchor frame plus $W$ previous 4-frame latent groups before the target group. Thus $W=0$ encodes frames [76, 81), while larger $W$ values include progressively more previous latent groups.

\begin{figure}[t]
\centering
\begin{minipage}[c]{0.48\linewidth}
\centering
\includegraphics[width=\linewidth]{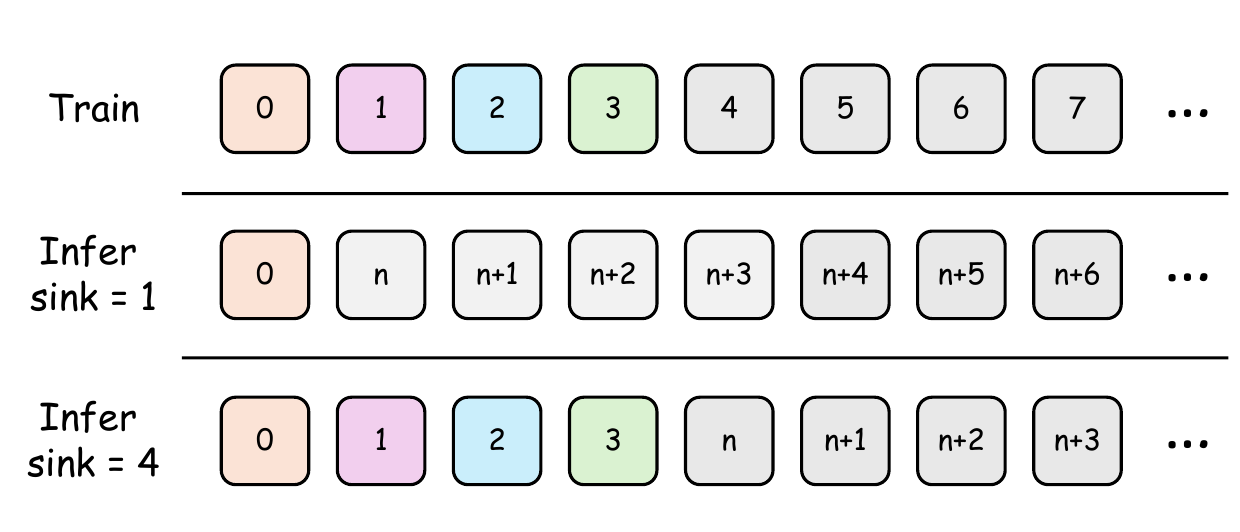}
\end{minipage}
\hfill
\begin{minipage}[c]{0.44\linewidth}
\centering
\includegraphics[width=\linewidth]{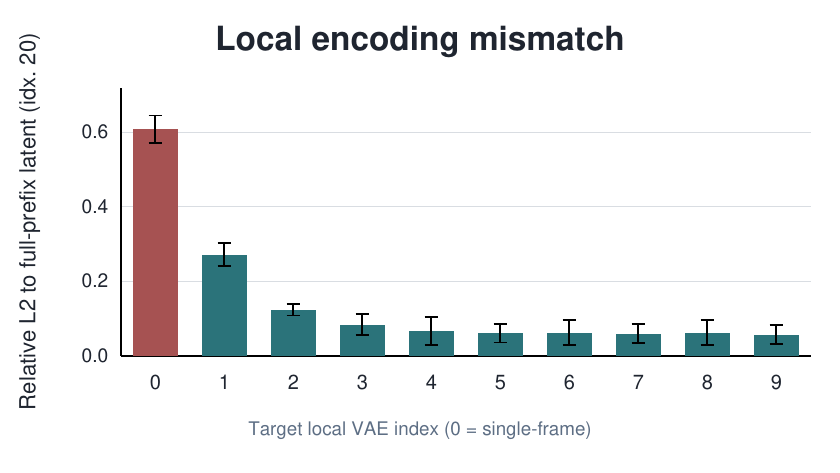}
\end{minipage}
\caption{\textbf{VAE boundary and sink-latent choice.} Left: sink 1 preserves only the stream-start anchor, while sink 4 preserves the short boundary-transition prefix. Right: local re-encoding mismatch to the full-prefix 21st latent drops sharply after adding previous latent groups and plateaus around the first four positions. This motivates using four sink latents for frame-wise streaming inference.}
\label{fig:vae-boundary-method}
\end{figure}

The diagnostic shows a large mismatch when the target group is encoded as a fresh stream: the relative L2 error is 0.607. Adding the boundary anchor reduces the error to 0.271 at $W=0$, and including more previous latent groups further reduces it to 0.124, 0.084, 0.067, and 0.060 for $W=1,2,3,4$, respectively. Beyond this range, the error changes little. This indicates that the early VAE latents form a short boundary-transition region rather than a single isolated sink token.

\paragraph{Sink-number ablation.}
We therefore ablate the number of sink latents in frame-wise SGF with TF initialization at 60 seconds. The total streaming context budget is fixed, so increasing the sink size preserves more prefix latents but leaves fewer slots for recent FIFO context. The relevant trend is therefore not strict monotonic improvement, but whether a small sink can cover the VAE boundary region without unnecessarily reducing the recent-context budget.

\begin{table}[t]
\centering
\caption{Sink ablation in frame-wise SGF with TF initialization on 60-second evaluation. Higher is better for all listed scores after VBench orientation.}
\label{tab:sink}
\small
\setlength{\tabcolsep}{5pt}
\begin{tabular*}{\linewidth}{@{\extracolsep{\fill}}lcccc}
\toprule
Metric & Sink 1 & Sink 2 & Sink 4 & Sink 8 \\
\midrule
Aesthetics & 0.627 & 0.645 & 0.653 & 0.655 \\
Background & 0.970 & 0.974 & 0.974 & 0.976 \\
Dynamics & 0.743 & 0.733 & 0.730 & 0.728 \\
Imaging & 0.715 & 0.713 & 0.714 & 0.714 \\
Motion & 0.982 & 0.982 & 0.982 & 0.983 \\
Subject & 0.983 & 0.982 & 0.983 & 0.983 \\
Flickering & 0.990 & 0.990 & 0.991 & 0.992 \\
\bottomrule
\end{tabular*}
\end{table}

Tab.~\ref{tab:sink} shows that sink 1 is weaker on aesthetic quality and flickering, while sink 4 reaches the stable range suggested by the VAE boundary diagnostic. Sink 8 gives only marginal additional gains on some metrics and consumes more of the fixed context budget. We therefore use sink 4 in the frame-wise long-horizon experiments as the smallest sink size that covers the observed boundary-transition prefix.

\begin{figure}[t]
\centering
\includegraphics[width=\linewidth]{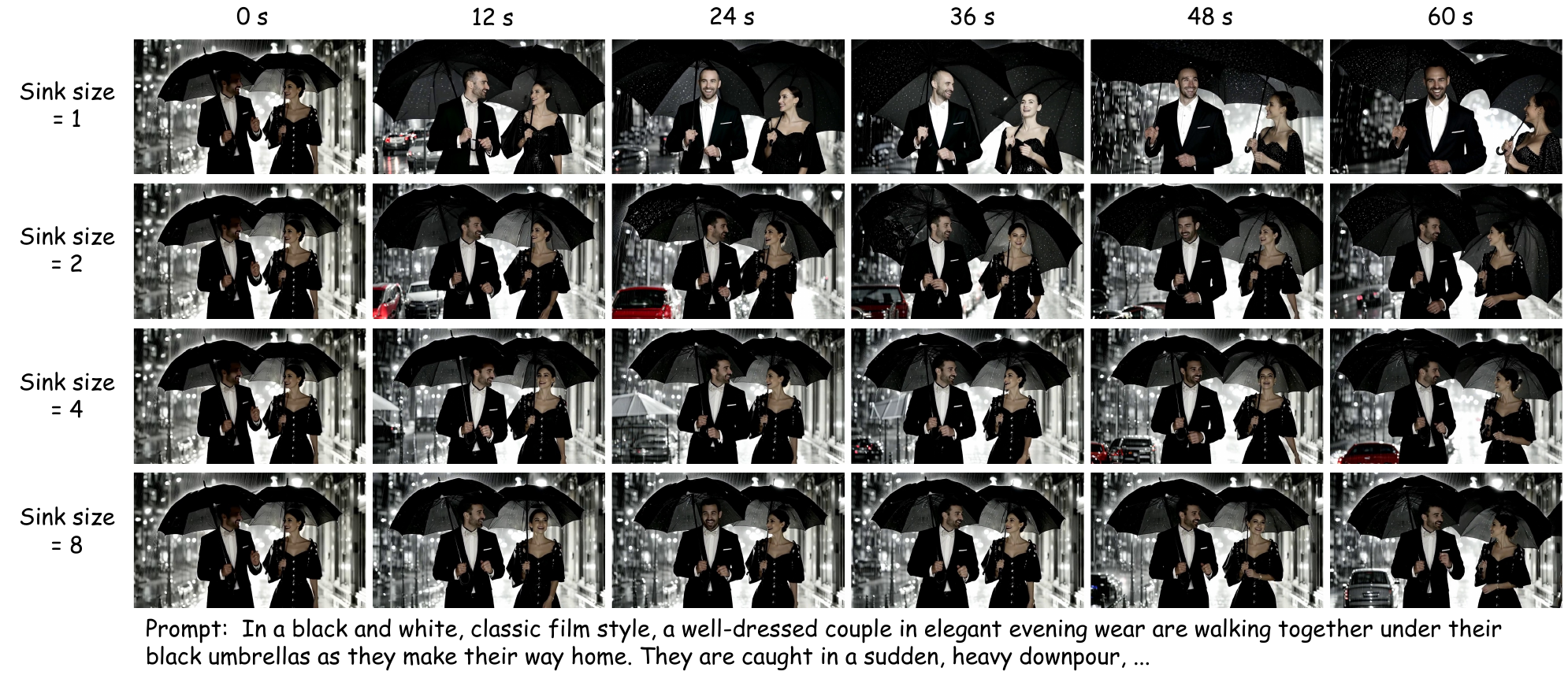}
\caption{\textbf{Qualitative sink ablation at 60 seconds.} Very small sinks preserve too little early stream-boundary information, while a multi-latent sink better maintains subject and scene consistency during long extrapolation. The ablation supports the selected frame-wise streaming policy; the core SGF claim remains the recovery of gradients through self-generated context K/V formation.}
\label{fig:sink-qual}
\end{figure}
\section{Additional Related-Work Discussion}
\label{sec:additional-related-work-discussion}

During revision, we recognized that Checkpointed Self Forcing in Solaris~\citep{solaris} is closely related to SGF, and include this discussion to clarify the connection. Solaris introduces Checkpointed Self Forcing for multiplayer Minecraft world modeling. In their TPU-based implementation, it performs a no-gradient autoregressive rollout to cache clean estimates and noisy transition states, then replays the sampled denoising step in parallel under a teacher-forcing attention mask. This reduces the memory cost of sliding-window Self Forcing and makes gradients through recomputed K/V representations feasible.

SGF uses a similar rollout-then-parallel-recompute structure, but formulates it as context-gradient reconstruction for the historical context-gradient gap. In frozen-cache self-rollout training, future losses supervise how noisy target tokens read cached history, but not how earlier self-generated latents are written into future-readable K/V memory. SGF keeps the recorded generated latents stop-gradient and does not backpropagate through the serial rollout; instead, future video-latent losses supervise the recomputed clean-context K/V writer under the matched causal mask. Our experiments study this bounded K/V-gradient recovery objective for native long-video extrapolation across frame-wise and chunk-wise generation, multiple initializations, and 5s/60s/240s horizons.

\section{Limitations}
\label{sec:limitations}

\methodshort{} is a bounded surrogate for a missing context-gradient signal rather than full backpropagation through the autoregressive rollout. It supervises how recorded self-generated latents are written into future-readable memory, but it does not update the sampled latents themselves through future losses, nor does it optimize the sequence of denoising decisions that produced them. We therefore do not claim to recover the exact gradient of the full serial rollout.

\methodshort{} also assumes that the parallel Pass-2 reconstruction faithfully reproduces the serial context relation at the sampled exit step. If the teacher-forcing mask, sink positions, FIFO window, RoPE handling, context timestep, or chunk alignment deviates from inference, SGF may recover gradients for the wrong attention relation and thus train a different writer. This alignment is especially important at the clean context timestep, precisely because that cache-writing call is shared with, but not directly supervised by, the noisy exit-step losses in frozen-cache training.

Finally, SGF is not a replacement for other long-video techniques. Streaming long-rollout tuning, retrieval-augmented memory, sparse attention, stronger causal initialization, and long-context teachers all target complementary parts of the system. Our claim is narrower: self-rollout training uses generated history as context but leaves a specific history-formation gradient unused, and SGF offers a practical way to recover it. A natural next step is to combine SGF with long-rollout exposure or retrieval, so that the model both writes better short-window memory and can access richer long-range context.

\clearpage
\section{Additional Qualitative Results}
\label{sec:additional-qual}

This appendix provides additional long-horizon qualitative comparisons for the frame-wise and chunk-wise experiments. The strips are intended to complement the quantitative tables by showing the temporal failure modes behind the aggregate scores. Each comparison should be read within a matched setting: Self Forcing and \methodshort{} use the same prompt, seed, initialization, horizon, sampling configuration, and inference context geometry. We focus on memory-related drift, including view changes, crop drift, scene replacement, subject identity changes, object disappearance, and collapse into unrelated textures.

\subsection{Frame-wise Comparisons}

Frame-wise generation writes each generated latent frame back into the historical context, so errors in historical K/V formation can accumulate at the finest temporal granularity. Across TF, causal CD, and causal ODE initializations, Self Forcing often remains locally plausible in early frames but gradually changes the subject, camera distance, object layout, or background. Under weaker initializations, the drift can become severe, producing cropped fragments, color blocks, or unrelated scene textures. \methodshort{} consistently reduces these failure modes: it better preserves the subject-scene relation, camera framing, and object layout over both 60-second and 240-second rollouts. 

\begin{figure}[t]
\centering
\includegraphics[width=\linewidth]{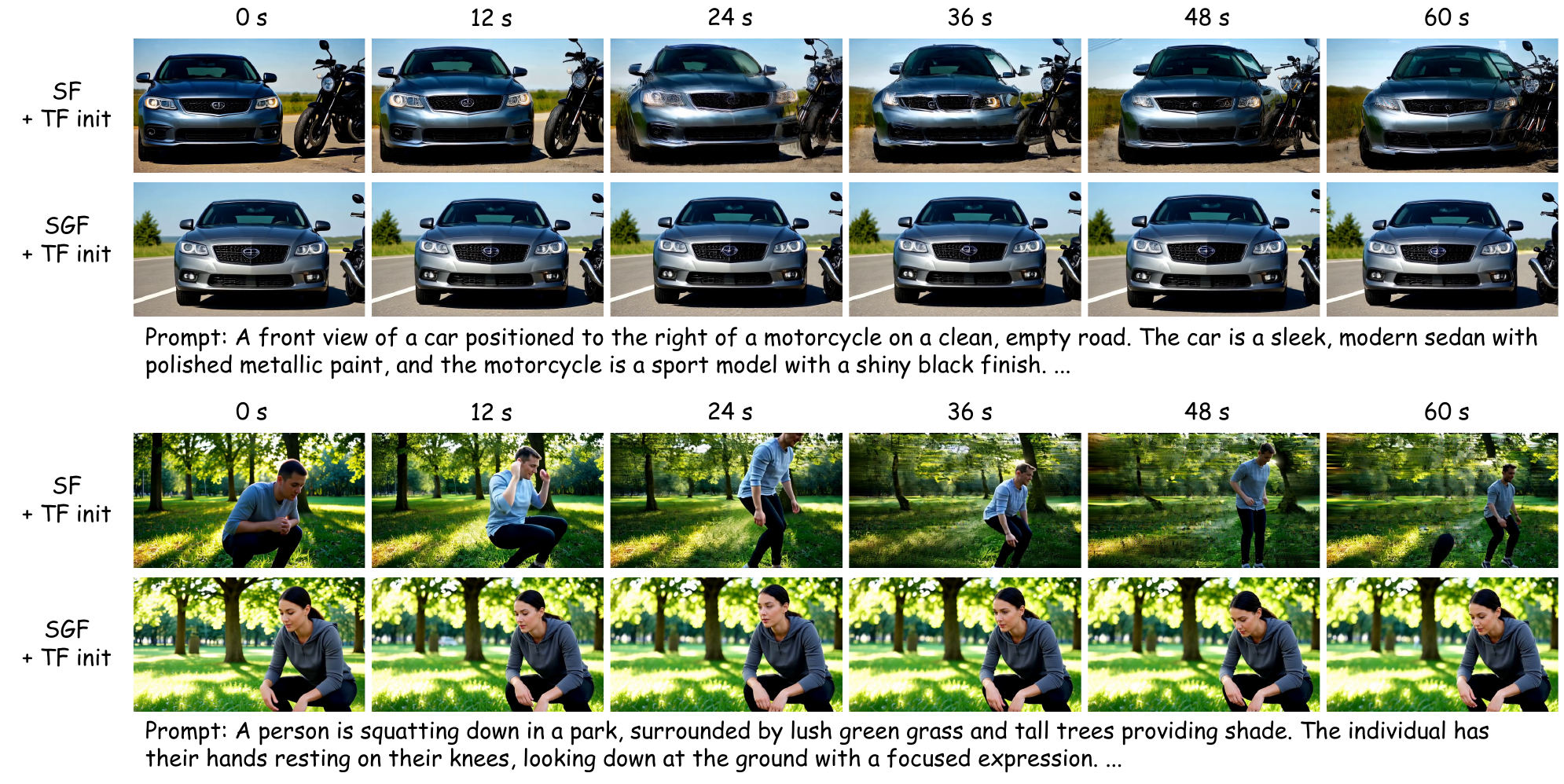}
\caption{\textbf{Frame-wise 60-second comparison under TF initialization.} In the car-and-motorcycle prompt, both methods remain plausible, but SGF keeps the car position, road geometry, and motorcycle relation more fixed. In the squatting-person prompt, Self Forcing changes the subject pose, identity, and framing over time, whereas SGF maintains a consistent squatting action and park background. This example illustrates that SGF can reduce identity and camera drift even before catastrophic visual breakdown.}
\label{fig:framewise-tf-60-main}
\end{figure}


\begin{figure}[t]
\centering
\includegraphics[width=\linewidth]{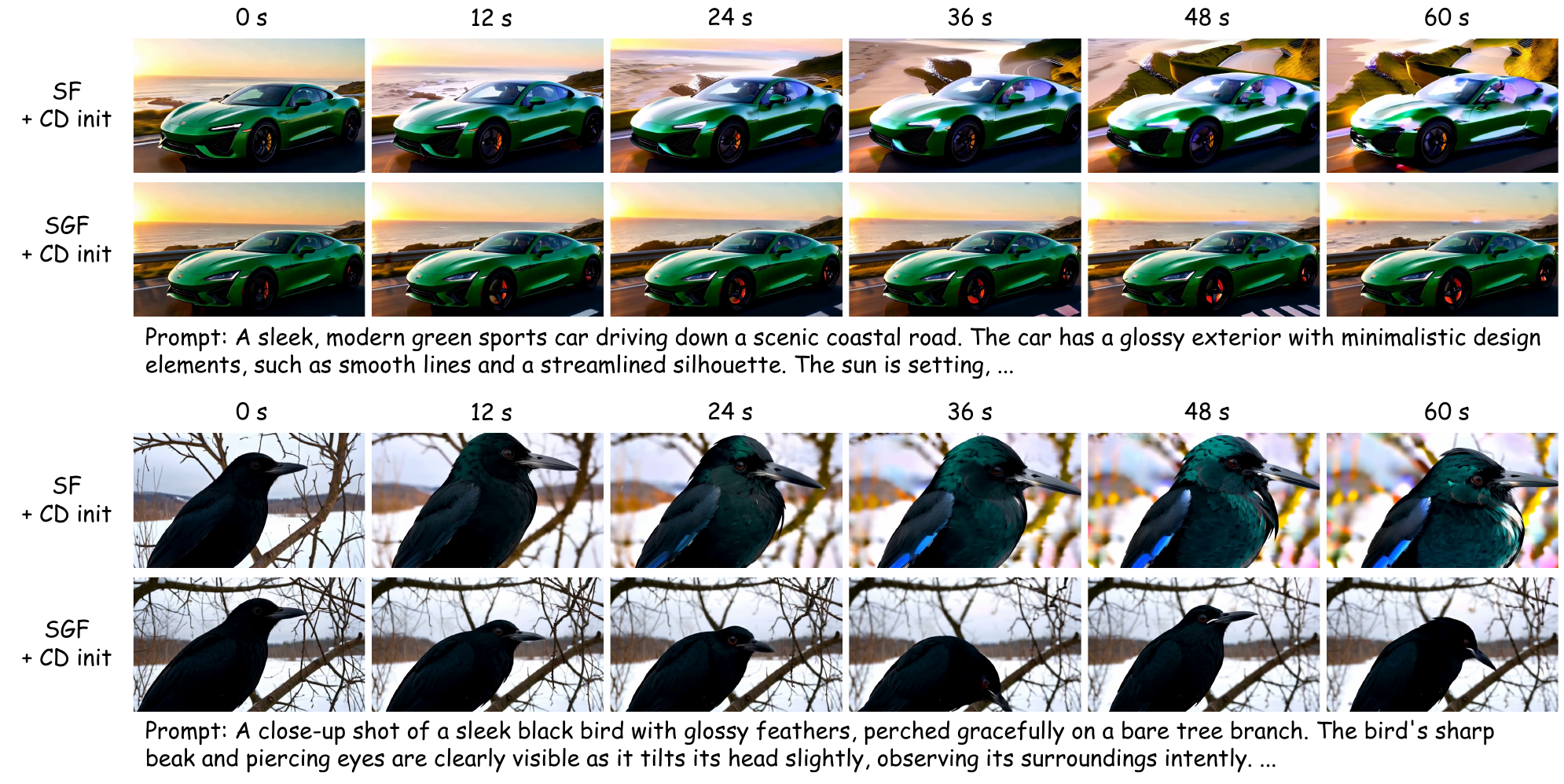}
\caption{\textbf{Frame-wise 60-second comparison under causal CD initialization.} Under causal CD initialization, Self Forcing does not always collapse, but it exhibits systematic crop and scale drift. The sea-turtle and elephant examples progressively zoom into the subject, reducing scene context and sometimes clipping the object. SGF keeps the animal scale, background, and camera distance more stable, showing that SGF improves non-catastrophic long-horizon composition drift as well as outright failure.}
\label{fig:framewise-cd-60}
\end{figure}

\begin{figure}[t]
\centering
\includegraphics[width=\linewidth]{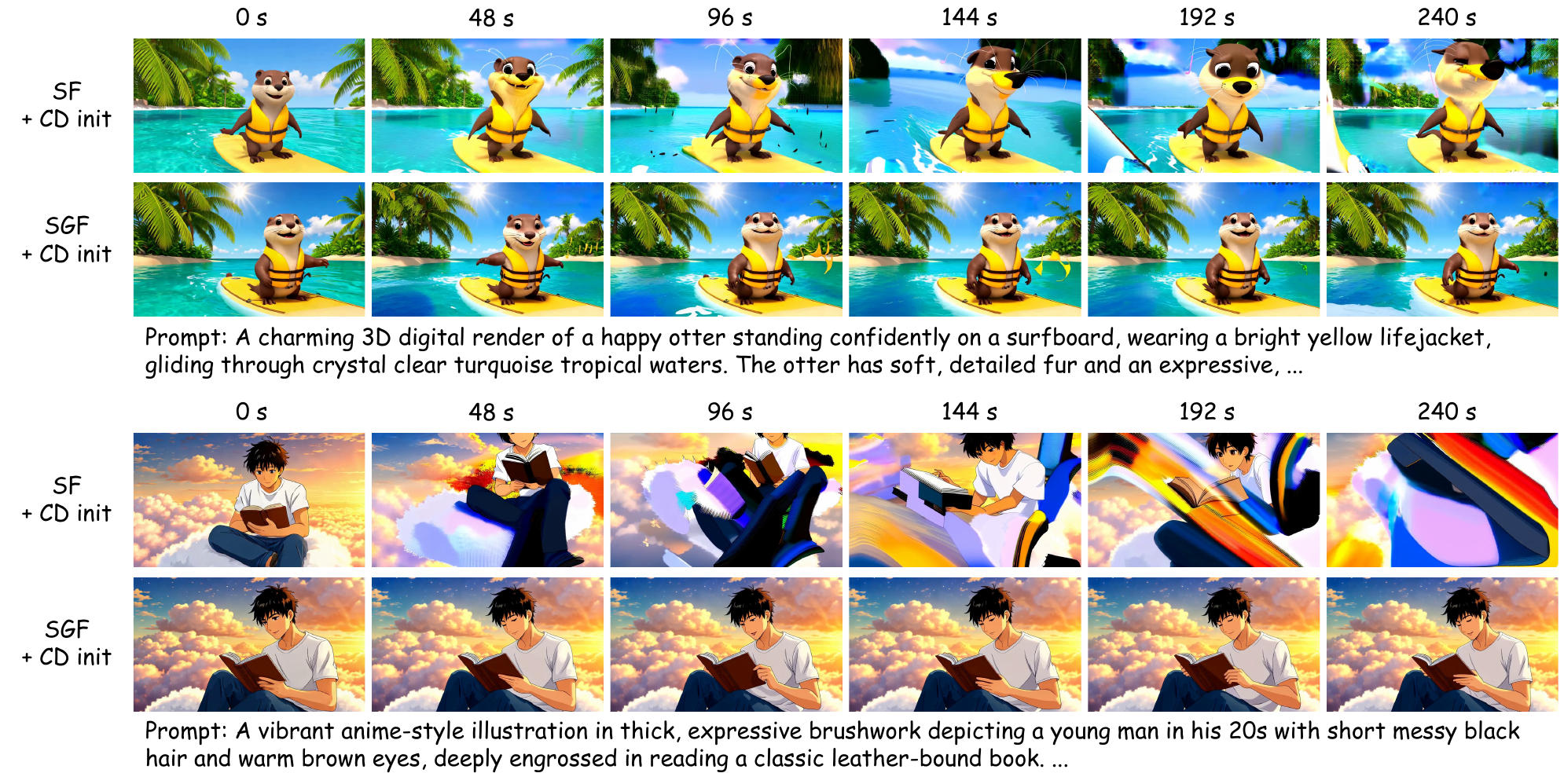}
\caption{\textbf{Frame-wise 240-second comparison under causal CD initialization.} The otter prompt shows that Self Forcing can preserve broad semantics but accumulates pose and boundary distortions, while SGF keeps the otter centered on the surfboard with a more stable tropical-water layout. In the anime reading prompt, Self Forcing drifts into partial crops and high-saturation color streaks by the late timestamps; SGF preserves the boy, book, and cloud background over the full 240 seconds.}
\label{fig:framewise-cd-240}
\end{figure}

\begin{figure}[t]
\centering
\includegraphics[width=\linewidth]{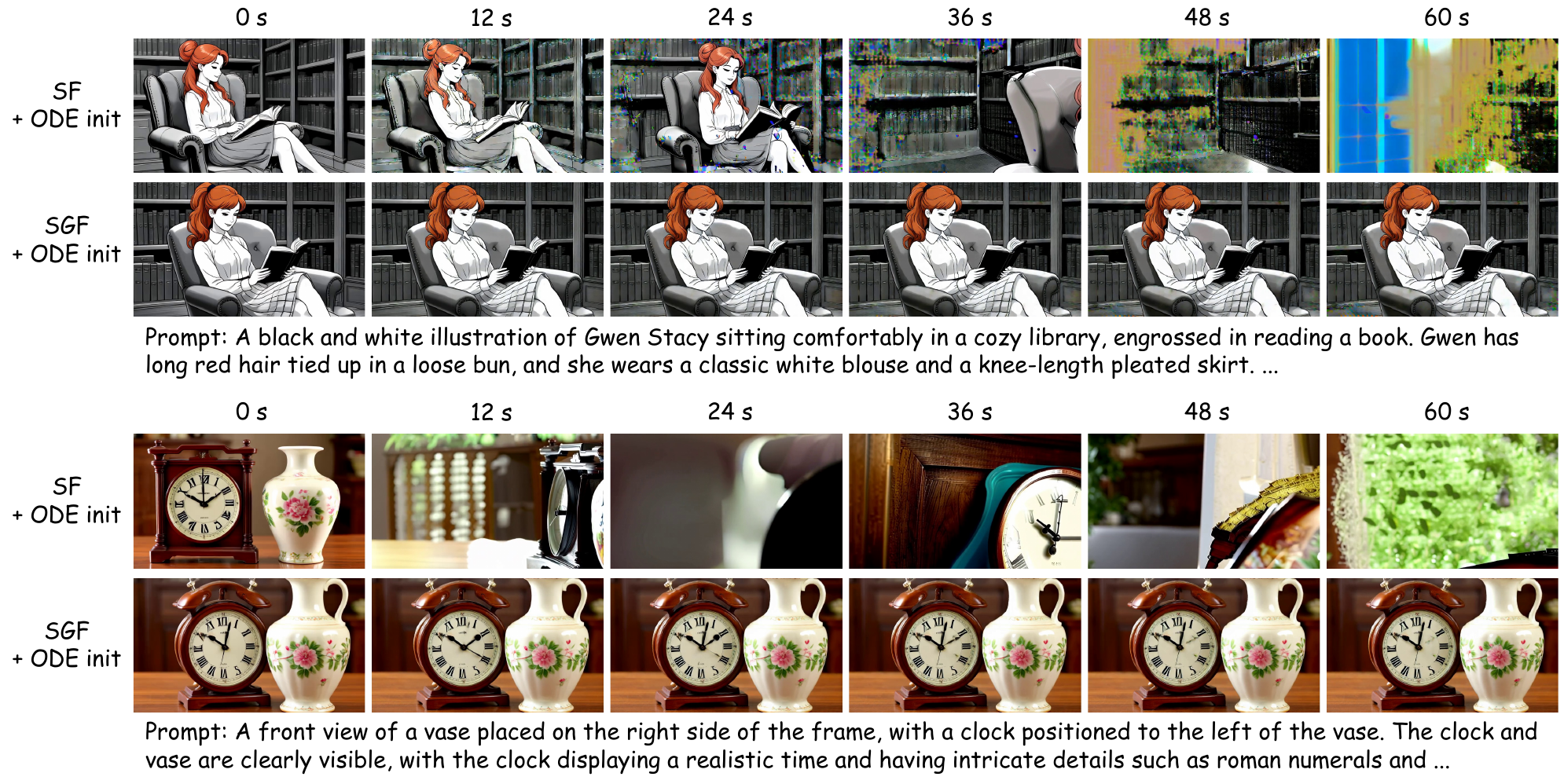}
\caption{\textbf{Frame-wise 60-second comparison under causal ODE initialization.} In the library-reading prompt, Self Forcing introduces chromatic background artifacts by 24 seconds, drops the seated reader by 36 seconds, and devolves into bookshelf/window fragments by 48--60 seconds; SGF keeps the reader, armchair, open book, and bookcase layout stable. In the clock-and-vase prompt, Self Forcing drifts from the intended frontal two-object composition into cropped or blurred close-ups and background fragments, whereas SGF preserves the clock-left/vase-right tabletop arrangement across the rollout.}
\label{fig:framewise-ode-60}
\end{figure}

\begin{figure}[t]
\centering
\includegraphics[width=\linewidth]{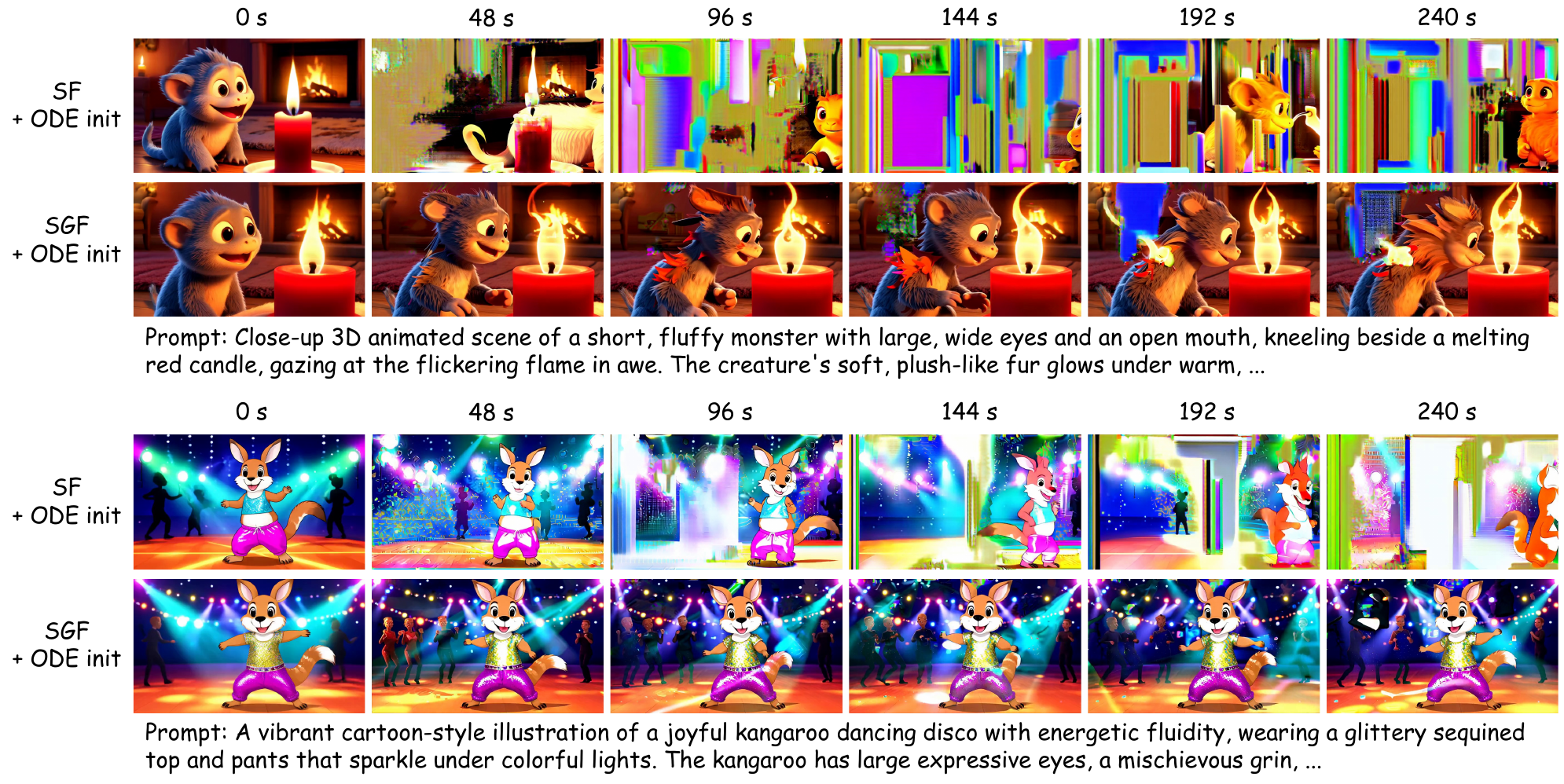}
\caption{\textbf{Frame-wise 240-second comparison under causal ODE initialization.} Under matched initialization and frame-wise streaming settings, Self Forcing undergoes severe long-horizon drift: the candle/creature example degenerates into color blocks and partial subject replacements, and the disco kangaroo example repeatedly loses the stage layout and subject identity. SGF preserves the prompt-specific subject, pose family, and scene layout more consistently through 240 seconds.}
\label{fig:framewise-ode-240}
\end{figure}

\clearpage
\subsection{Chunk-wise Comparisons}

Chunk-wise generation updates memory at a coarser temporal granularity, with each generated chunk serving as context for later chunks. The chunk-wise figures test whether the same recovered context-gradient signal remains useful when history is written and consumed in multi-frame blocks. When ODE-initialized rows are included, they are reference baselines; the controlled comparisons are the matched Self Forcing and \methodshort{} rows under the same initialization. The qualitative pattern is consistent with the frame-wise case: Self Forcing can replace actions, change people, lose objects, or drift into unrelated textures, while \methodshort{} better maintains the action, subject identity, and scene layout over long rollouts. 

\begin{figure}[t]
\centering
\includegraphics[width=\linewidth]{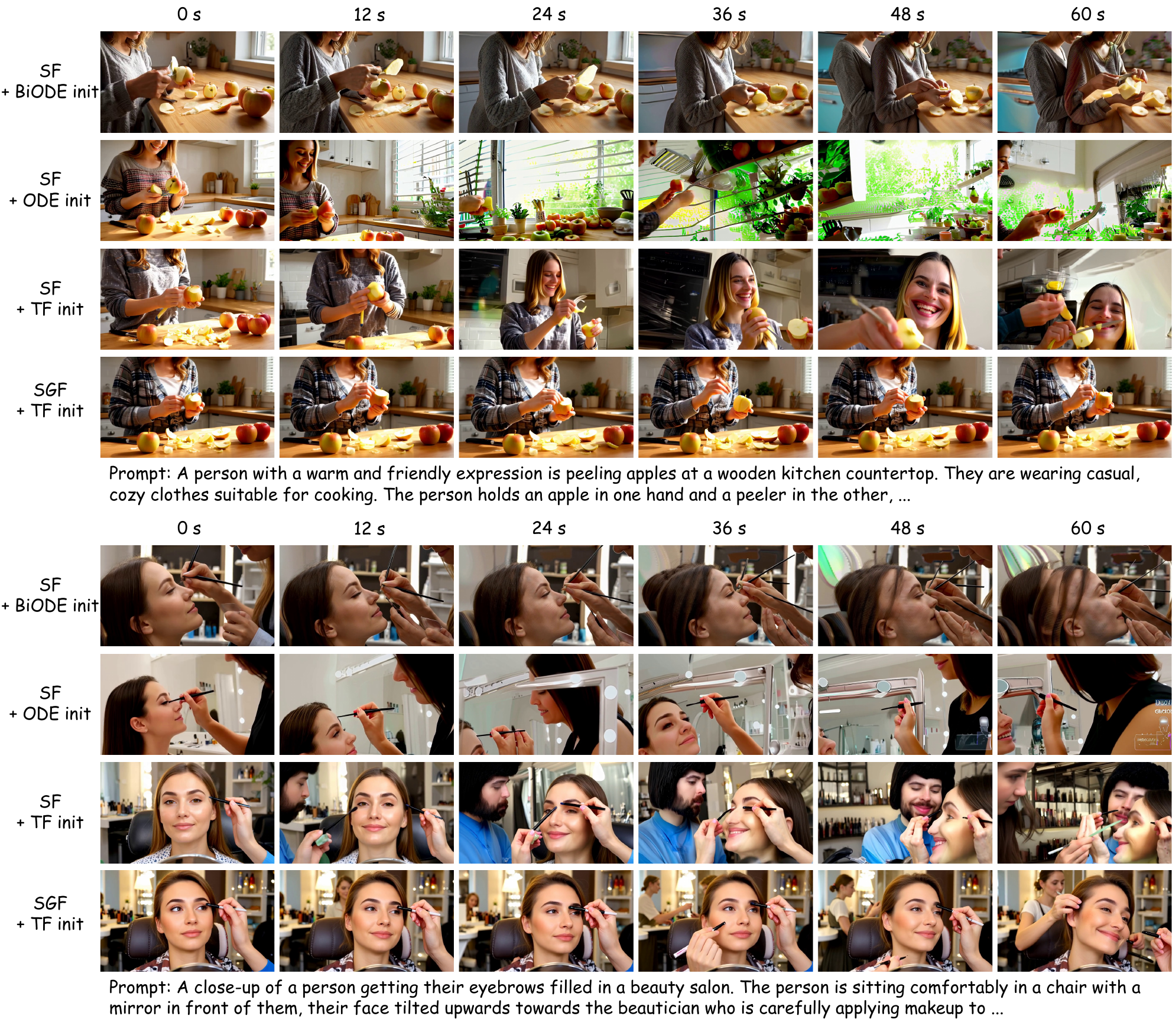}
\caption{\textbf{Chunk-wise 60-second comparison under TF initialization.} In the apple-peeling prompt, Self Forcing with TF initialization changes the person identity and camera viewpoint, at times replacing the tabletop action with close-ups of a smiling face. SGF keeps the hands, apples, and countertop action more consistent. In the eyebrow-makeup prompt, Self Forcing changes the number and identity of people across time; SGF maintains the same seated client, makeup gesture, and salon environment.}
\label{fig:chunkwise-tf-60}
\end{figure}

\begin{figure}[t]
\centering
\includegraphics[width=\linewidth]{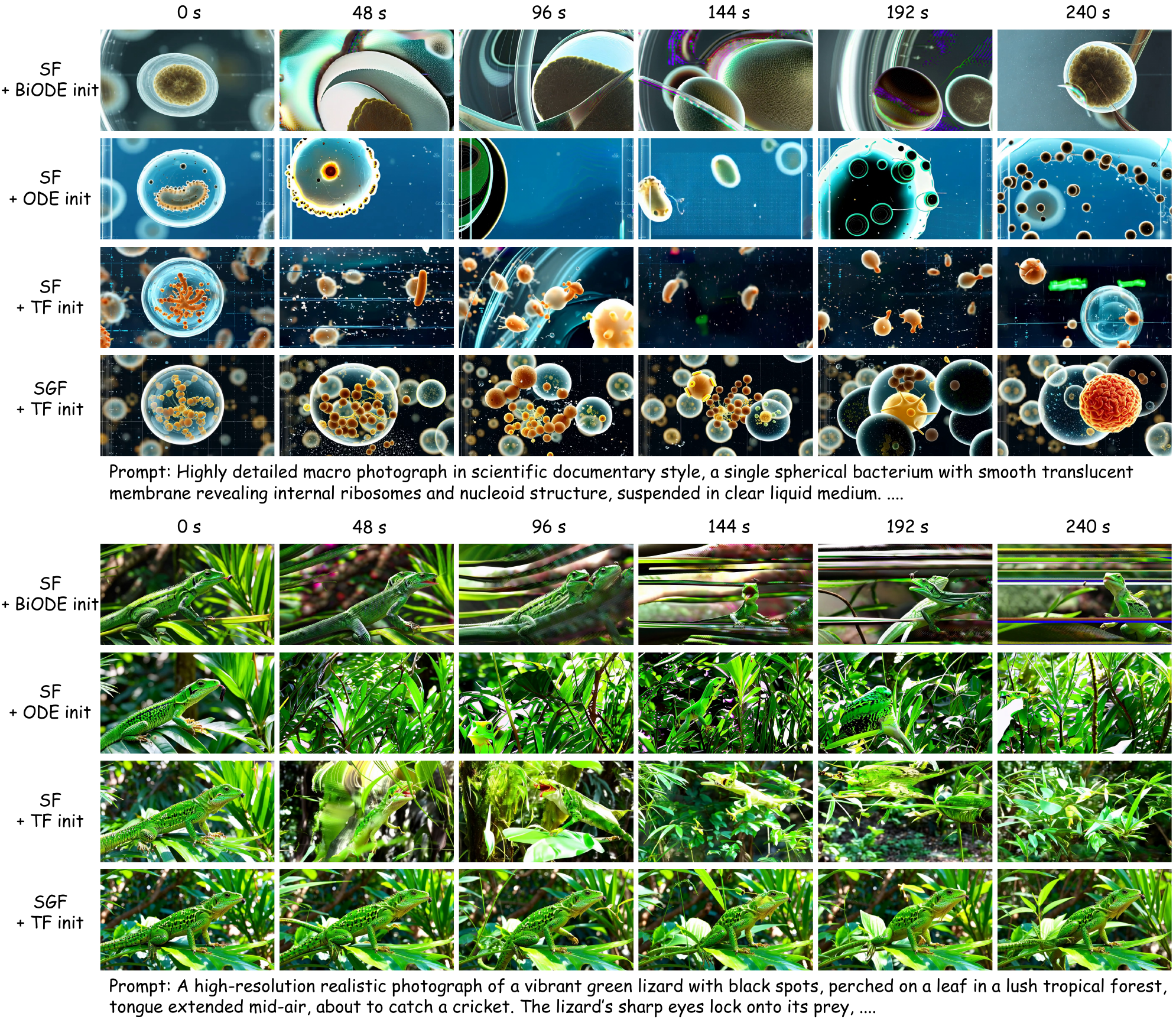}
\caption{\textbf{Chunk-wise 240-second comparison under TF initialization.} The coarser chunk-wise rollout tests whether a generated block remains useful as context several chunks later. In the bacterium prompt, SGF maintains a population of translucent spherical cells more consistently, whereas Self Forcing often turns the scene into unrelated circular textures or sparse particles. In the lizard prompt, Self Forcing frequently loses the animal or collapses into foliage; SGF better preserves a green lizard among leaves across the long horizon.}
\label{fig:chunkwise-tf-240}
\end{figure}

\begin{figure}[t]
\centering
\includegraphics[width=\linewidth]{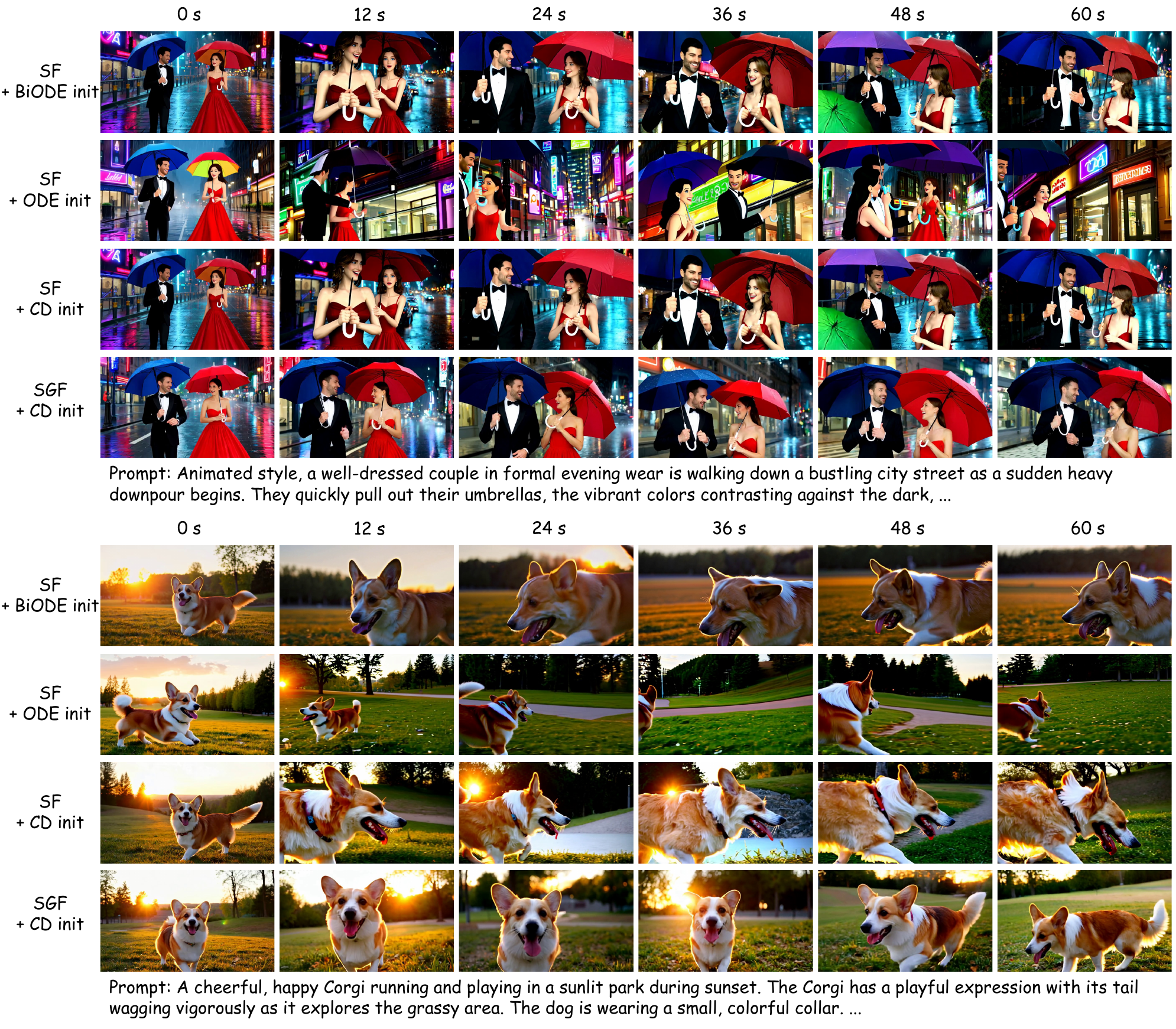}
\caption{\textbf{Chunk-wise 60-second comparison under causal CD initialization.} The strip includes bidirectional-ODE and causal-ODE Self Forcing references in addition to the matched causal-CD pair. In the umbrella-couple prompt, SGF under causal CD initialization keeps two people, umbrella colors, and rainy street layout more coherent, while Self Forcing variants often change viewpoint or subject arrangement. In the corgi prompt, SGF under causal CD initialization preserves a visible running dog and grassy park context more consistently than the matched Self Forcing row, which drifts toward close-up crops.}
\label{fig:chunkwise-cd-60}
\end{figure}

\begin{figure}[t]
\centering
\includegraphics[width=\linewidth]{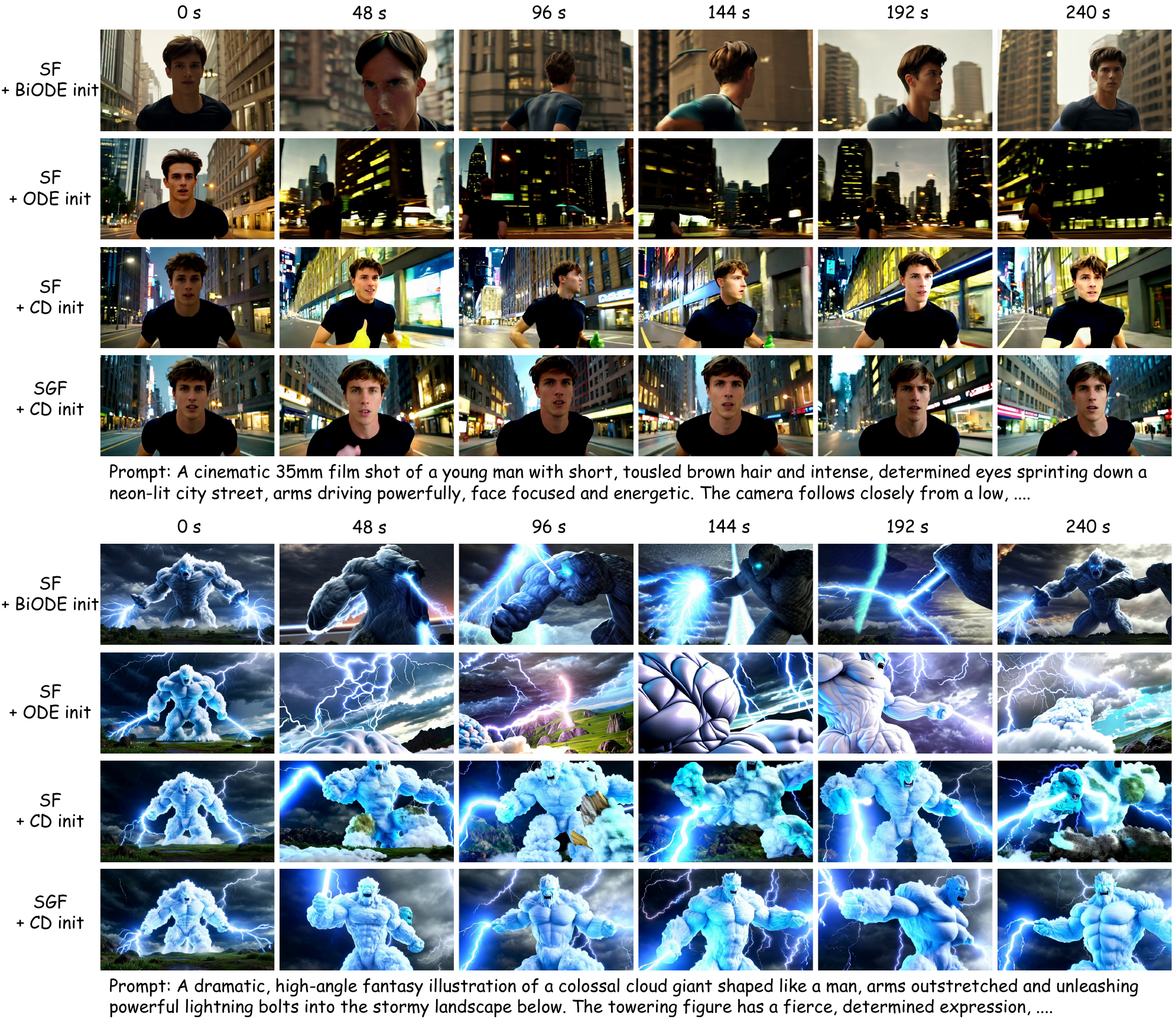}
\caption{\textbf{Chunk-wise 240-second comparison under causal CD initialization.} This strip includes ODE-initialized Self Forcing references and a matched causal-CD pair. In the running-man prompt, the ODE baselines drift into night streets or back-view shots, while matched Self Forcing under causal CD initialization preserves the subject intermittently but still changes viewpoint and facial identity. SGF under causal CD initialization keeps a front-facing runner and neon street context more consistently. In the cloud-giant prompt, SGF better maintains a coherent humanoid lightning figure instead of collapsing into close-up fragments or unrelated storm textures.}
\label{fig:chunkwise-cd-240}
\end{figure}

\end{document}